\documentclass[10pt,twocolumn,letterpaper]{article}

\usepackage[pagenumbers]{wacv} 

\definecolor{wacvblue}{rgb}{0.21,0.49,0.74}
\usepackage[pagebackref,breaklinks,colorlinks,allcolors=wacvblue]{hyperref}

\def\wacvPaperID{103} 
\def\confName{WACV}
\def\confYear{2027}

\usepackage[utf8]{inputenc} 
\usepackage[T1]{fontenc}    
\usepackage{url}            
\usepackage{booktabs}       
\usepackage{amsfonts}       
\usepackage{nicefrac}       
\usepackage{microtype}      
\usepackage[dvipsnames]{xcolor}     

\usepackage{graphicx}
\usepackage{multirow}
\usepackage{adjustbox}
\usepackage{tabularx}
\usepackage{xspace}
\usepackage[normalem]{ulem}
\usepackage{tcolorbox}
\usepackage{mdframed}
\usepackage{courier} 
\usepackage{makecell}
\usepackage{amsmath}
\usepackage{subcaption}

\useunder{\uline}{\ul}{}

\newcommand{\methodname}{FiGMo\xspace}

\title{Fine-grained Human Motion Understanding with Language Models}

\author{Thomas Markhorst$^1$, Zhi-Yi Lin$^1$, Jouh Yeong Chew$^2$, Jan van Gemert$^1$, Xucong Zhang$^1$\\
$^1$Delft University of Technology, $^2$Honda Research Institute Japan\\
{\tt\small t.c.markhorst@tudelft.nl}
}

\begin{document}
\maketitle
\begin{abstract}
Fine-grained motion understanding requires knowing not only which poses occur, but also when they occur. Existing motion-language models either compress motion clips or concatenate pose features without retaining the timing of each pose, making action duration and temporal scale ambiguous under temporal subsampling. We introduce FiGMo, an LLM-based model that represents motion as timestamp-grounded skeletal poses, supporting reasoning over action order and duration. A unified pose encoder applies this representation to both 2D and 3D skeletons, including poses detected from real videos, while RGB can optionally provide contextual information. To reason over timestamped pose sequences, the model must understand both individual body configurations and how they change over time. As existing data provides limited pose-level grounding, we construct the required pose-to-motion supervision. Controlled experiments show that timestamps improve temporal localization and preserve physical timing under subsampling, while the constructed supervision drives the main performance gains. FiGMo achieves state-of-the-art results across motion question answering, dense motion captioning, and action recognition with detected poses. Our code is available at: \href{https://github.com/Thomas-Markhorst/FiGMo}{https://github.com/Thomas-Markhorst/FiGMo}
\end{abstract}      
\section{Introduction}
\label{sec:introduction}

Understanding human motion is a fundamental step toward building human-centered AI systems that can perceive, interpret, and interact with people in socially appropriate ways~\cite{witzig2024multimodal,zhu2025react,Lin_2026_CVPR}. It supports a wide range of applications, including human-robot interaction~\cite{sciutti2018humanizing}, rehabilitation monitoring~\cite{rehabilitation, assisted_privacy_preserving}, and sports coaching~\cite{shao2020finegym}.  
Despite substantial progress, achieving fine-grained and temporally grounded motion understanding remains challenging~\cite{chen2021sportscap,wang2023fg,shao2020finegym}.

\begin{figure}[tb]
    \centering
    \includegraphics[width=\linewidth]{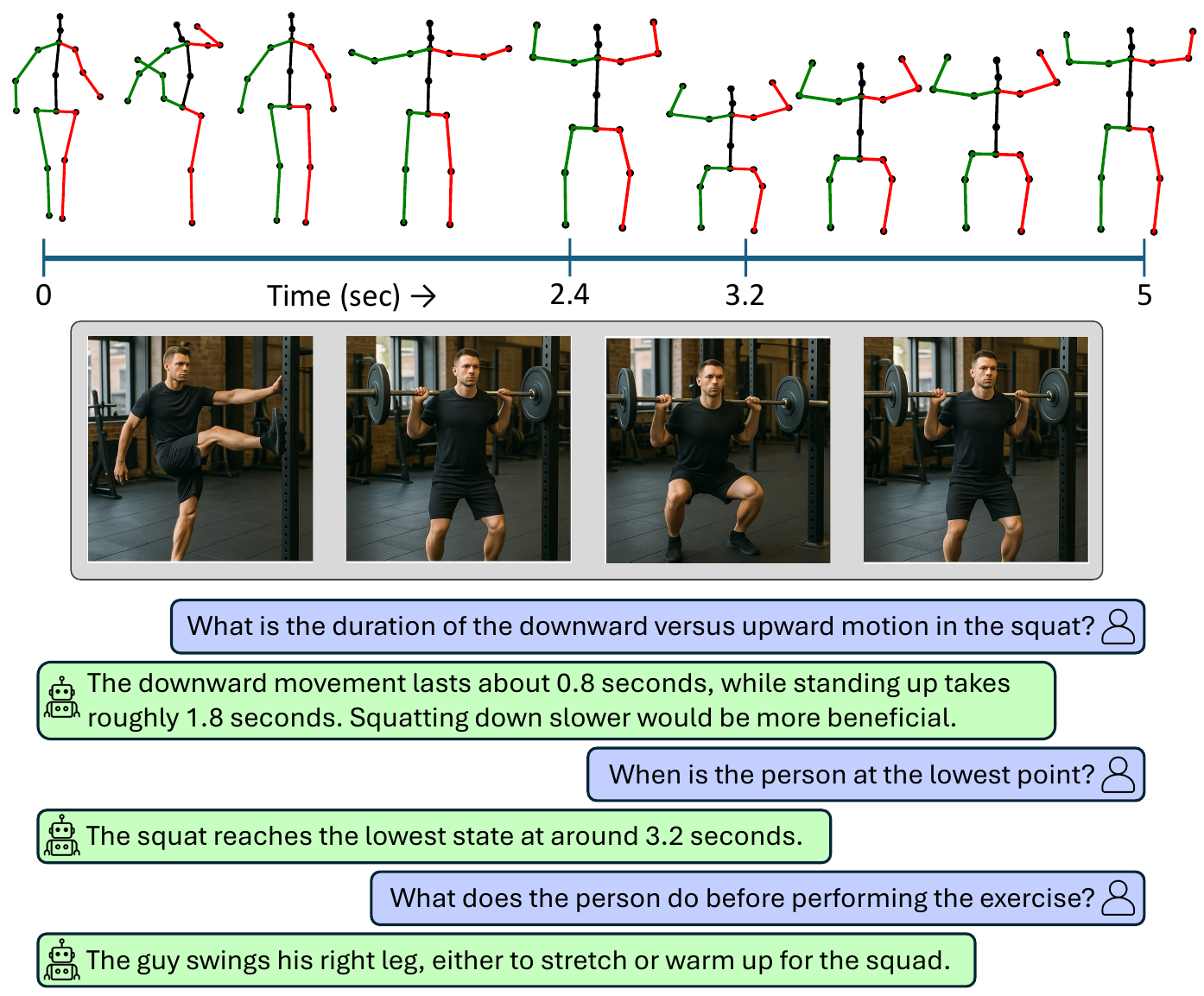}
    \caption{Given a motion sequence (top left), our method answers questions about fine-grained temporal aspects of human motion (right). By representing motion as timestamped skeletal poses, \methodname supports reasoning about action order, timing, and duration.
    }
    \label{fig:teaser}
\end{figure}

Recent advances in large language models (LLMs) have expanded multimodal reasoning across video understanding~\cite{lin2024video}, motion captioning~\cite{wu2024motion,guo2024momask}, and behavior analysis~\cite{chan2024human}. Most of these approaches rely on video as the primary input modality. While video provides rich visual context, it requires processing high-dimensional pixel inputs that are computationally expensive to process and therefore typically require aggressive temporal subsampling~\cite{li2024llavaonevision,zhang2025videollama3,li2024video}. This often limits temporal precision and makes it difficult to capture subtle, rapid, or fine-grained motion dynamics. Moreover, raw video inherently contains appearance information such as identity, clothing, and background context, which may introduce privacy concerns or demographic biases in downstream applications~\cite{assisted_privacy_preserving, mehrabi2022surveybiasfairnessmachine,gender_shades,baltaretu2025smplly}. 
Therefore, in many realistic deployment scenarios, bandwidth~\cite{edge_cloud_collaboration}, regulatory~\cite{GDPR2016a}, or privacy constraints~\cite{Karkazis02012017} may prevent storing or transmitting video, particularly on edge devices~\cite{edge_cloud_computing}.

In contrast, representing humans as sequences of body joints provides a compact, explicit, and appearance-invariant description of motion~\cite{privacy_preserving,duan2022revisiting,muppet,zhu2023motionbert}. Such skeletal representations are significantly lower-dimensional than raw video and therefore enable high frame-rate processing while abstracting away sensitive visual attributes~\cite{privacy_preserving}. This makes skeleton-based reasoning attractive for privacy-sensitive, fine-grained applications such as rehabilitation monitoring. 

However, existing human motion-centric models often overlook detailed temporal reasoning. Several motion-language models quantize motion into discrete latent codes~\cite{fang2025humocon,jiang2023motiongpt,wang2024motiongpt2}, which can obscure the correspondence between individual poses and physical time. Other approaches retain continuous pose features~\cite{li2025human}, but do not explicitly represent the temporal origin of each pose. Consequently, motion clips are treated as atomic units rather than temporally structured sequences, limiting the model's ability to reason about the exact timing and duration of motion.

In this work, we present \methodname, an LLM-based method for \underline{\textbf{Fi}}ne-\underline{\textbf{G}}rained \underline{\textbf{Mo}}tion understanding that models human motion as a temporally grounded sequence of skeletal poses. Each pose is associated with a timestamp, preserving its temporal origin within the motion sequence. To reason over these timestamped sequences, the model must understand both individual body configurations and how they change over time. As data provides limited pose-level grounding, we construct the required pose-to-motion supervision, spanning pose and motion captioning and question answering. We compare training on the complete mixture in a single stage with a staged pose-to-motion schedule; our ablations show that supervision diversity drives the main gains, while staging provides a smaller additional benefit.

Unlike many prior motion-based methods that rely on clean 3D MoCap skeletons~\cite{fang2025humocon,chen2024motionllm,feng2024chatpose} captured in controlled environments~\cite{mahmood2019amass}, our approach supports both 2D and 3D skeleton inputs. We train a shared pose encoder on mixed and partially corrupted 2D and 3D skeletal data, improving robustness to noisy 2D pose detections from monocular videos. When available, our method can incorporate video to provide supplementary context.

Across motion question answering, dense motion captioning, and detected-pose action recognition, \methodname achieves state-of-the-art performance, demonstrating the effectiveness of explicit timestamp grounding, unified 2D/3D pose encoding, and diverse pose- and motion-level supervision.

In summary, our contributions are:
\begin{itemize}
    \item per-pose timestamp grounding using a shared 2D/3D pose encoder, preserving temporal scale under subsampling;

    \item a unified 2D/3D motion-language system trained on newly constructed pose- and motion-level captioning and QA supervision, supporting detected poses and optional RGB;

    \item analyses confirming that timestamp grounding preserves temporal scale and the proposed supervision drives performance, with state-of-the-art results across benchmarks.

\end{itemize}

\section{Related Work}
\subsection{Motion Representation}
Early action recognition models focused on end-to-end video classification using convolutional and transformer-based architectures.  
Two-dimensional CNNs such as TRN~\cite{zhou2018temporal} and TSM~\cite{lin2019tsm} captured short-term temporal cues through sparse frame sampling, while 3D CNNs, including C3D~\cite{tran2015learning} and I3D~\cite{carreira2017quo} extended these ideas to dense spatio-temporal modeling. More recently, Video-LLMs have demonstrated strong performance in appearance-based understanding tasks~\cite{tang2025videounderstandinglargelanguage, yang2023vid2seqlargescalepretrainingvisual}. However, they operate on high-dimensional pixel inputs that often require aggressive temporal subsampling to remain computationally feasible~\cite{li2024llavaonevision, zhang2025videollama3}. 
This can obscure subtle motion cues and temporal relationships. 
Furthermore, video contains appearance information such as identity and ethnicity, which may introduce privacy or deployment constraints in real-world applications~\cite{assisted_privacy_preserving, mehrabi2022surveybiasfairnessmachine,gender_shades,baltaretu2025smplly}.

Skeleton-based approaches alleviate these issues by modeling motion as joint coordinate sequences rather than raw appearance.  
Graph-based and transformer-based architectures such as ST-GCN~\cite{yan2018spatial}, PoseC3D~\cite{duan2022revisiting}, and MotionBERT~\cite{zhu2023motionbert} are robust to viewpoint changes and support high-frame-rate processing.  
MotionBERT further improves generalization through AMASS pretraining~\cite{mahmood2019amass} with random joint masking and additive noise. However, these models remain confined to clip-level classification~\cite{carreira2017quo, hmdb, soomro2012ucf101dataset101human} without explicit reasoning about fine-grained motion timing~\cite{endo2023motion}.

\subsection{Human Motion Understanding}

To move beyond coarse skeletal action classification, several datasets have been introduced for more fine-grained motion understanding.  
HumanML3D~\cite{humanml3d}, BABEL~\cite{punnakkal2021babel}, CompMo~\cite{demo_3dv} and Motion-X~\cite{lin2023motionx, zhang2025motion} provide motion–language pairs with semantics and temporal annotations. BABEL-QA~\cite{endo2023motion} and HuMMan-QA~\cite{hummanQA} go beyond captioning by posing questions about specific events and relationships within a motion sequence, enabling evaluation of reasoning over both the content and temporal dynamics of human motion.

Using these datasets, cross-modal alignment models such as MotionCLIP~\cite{tevet2022motionclip} learn shared motion–text embeddings for retrieval and semantic organization.  
This idea was extended by multimodal LLMs including MotionGPT~\cite{jiang2023motiongpt}, MotionGPT2~\cite{wang2024motiongpt2}, and AvatarGPT~\cite{zhou2024avatargpt}, enabling text-to-motion synthesis, captioning, and motion-editing. DEMO~\cite{demo_3dv} and UniMotion~\cite{unimotion} move toward finer temporal modeling by segmenting and captioning motion sequences. Although capable of generating motion descriptions, these models perform captioning rather than answering questions about the content and temporal dynamics of the motion sequence.

Recent LLM-based models such as MotionLLM~\cite{chen2024motionllm} and HuMoCon~\cite{fang2025humocon} extend earlier captioning approaches by training on motion question–answering datasets~\cite{endo2023motion}. However, they largely retain the motion representations used for captioning. In particular, several works encode motion using discrete VQ-VAE tokens~\cite{fang2025humocon, chen2024motionllm, wang2024motiongpt2}, which compress entire motion sequences into latent motion codes. This compression removes the explicit correspondence between individual poses and their temporal origin, preventing precise timestamp conditioning and limiting fine-grained temporal modeling. While LLaMo~\cite{li2025human} represents motion as a sequence of skeletal poses rather than a compressed unit, it does not explicitly encode the temporal origin of each pose and instead focuses on key pose frames, making the temporal intervals between poses ambiguous. In contrast, our method represents motion as a sequence of skeletal poses with explicit timestamps, allowing the model to reason about motion timing at the pose level.

Another limitation of existing motion–LLM systems~\cite{chen2024motionllm, fang2025humocon, li2025human} is their reliance on clean 3D motion capture data collected in controlled environments~\cite{mahmood2019amass}. This restricts their applicability in real-world settings where motion is often obtained from noisy 2D pose detectors applied to monocular RGB videos. Therefore, our model applies a pose encoder pre-trained for both 2D and 3D poses.

Moreover, recent motion-LLM methods primarily adapt LLMs with motion-level QA supervision~\cite{fang2025humocon, chen2024motionllm, li2025human}. 
In contrast, several video-LLM training recipes first use image-language data or image-based instruction tuning before incorporating video-language data~\cite{li2024llavaonevision, zhang2025videollama3}. 
This suggests an analogous question for skeletal motion: whether pose-level supervision can support downstream motion understanding, and how it should be combined with motion-level supervision. We therefore construct a pose-to-motion supervision mixture and analyze how different components and training strategies affect skeletal motion-language reasoning.
\section{Method}
\label{sec:method}
\begin{figure*}[bt]
    \centering
    \includegraphics[width=\linewidth]{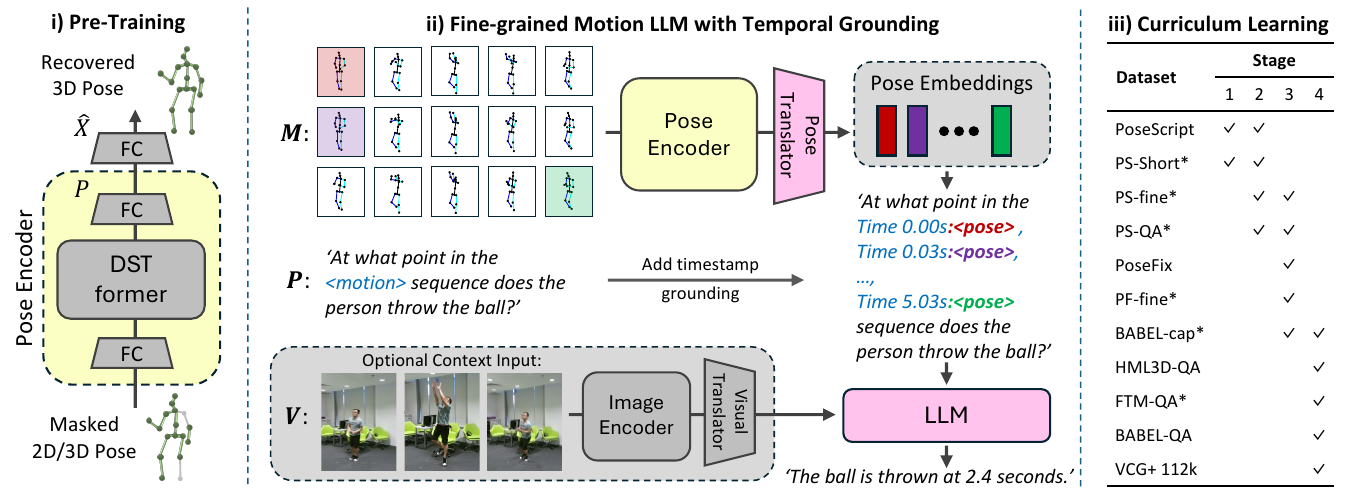}
    \caption{Pipeline of our \methodname, comprising (i) pose encoder pre-training, (ii) encoding high-frequency motion $M$ with explicit temporal grounding for fine-grained motion understanding, optionally with video input $V$ for contextual cues, and (iii) pose- and motion-level supervision modules that can be used either in a single stage or through a staged pose-to-motion schedule. Datasets marked with * are proposed in this work.}
    \label{fig:method}
\end{figure*}

We develop an LLM-based method for fine-grained human motion understanding that models human movement as temporally grounded skeletal pose sequences. The model accepts both 2D and 3D skeletal motion.

Each pose feature is grounded with an explicit timestamp, providing the LLM with access to the temporal origin of each body configuration. Beyond the motion representation itself, we construct a pose-to-motion supervision mixture that includes pose captioning, pose QA, motion captioning, and motion QA. This allows us to study how pose- and motion-level supervision contribute to skeletal motion-language reasoning, and how different training strategies affect performance. Fig.~\ref{fig:method} provides an overview of the model.

\subsection{Unified Pose Encoder}
To enable fine-grained 2D and 3D human motion understanding, we model human motion as a temporal sequence of static poses, where each pose is independently encoded to capture detailed spatial configurations of body joints. This design allows the model to focus on precise body structure at the frame level, while temporal reasoning is later handled by the LLM (see Sec.~\ref{subsec:meth_temp}).  

For pose encoding, we build upon the motion encoder of MotionBERT~\cite{zhu2023motionbert}, originally developed for 2D-to-3D human pose lifting. Since our goal for the encoder is to represent individual poses rather than continuous motion, we retain only the spatial attention blocks to encode body joints. 

To make the encoder robust to noisy or incomplete inputs, we pre-train it to reconstruct 3D poses from corrupted 2D or 3D pose inputs. During training, we apply both additive noise and random joint masking~\cite{devlin2019bert,bao2022beitbert,maskedautoencoders}, which regularizes the model and simulates real-world conditions such as occluded body parts or missing depth information. The encoder is trained using a mixture of detected 2D poses, 3D mocap data~\cite{mahmood2019amass}, and 2D projections of 3D mocap sequences, allowing it to generalize across pose representations.

During pre-training, the pose encoder maps each pose into a latent embedding $P$ and reconstructs a 3D pose $\hat{X}$ following~\cite{zhu2023motionbert}. For mocap data with available 3D ground truth, we minimize a reconstruction loss $\mathcal{L}_{3D}$, while for 2D pose inputs, we use a re-projection loss $\mathcal{L}_{\text{rep}}$:
\begin{equation}
    \mathcal{L}_{\text{3D}} = \sum_{j=1}^{J} \|\hat{X}_j - X_j\|_2, \quad
    \mathcal{L}_{\text{rep}} = \sum_{j=1}^{J} \delta_j \|\hat{x}_j - x_j\|_2,
\end{equation}
with $j$ for joint index, $\delta_j$ as joint visibility, $X$ as 3D pose, $x$ as 2D pose, and \textasciicircum{} indicating reconstructed pose.  
This pretraining yields a unified, robust 2D/3D pose encoder for downstream fine-grained motion reasoning.

\subsection{Multimodal LLM Structure}
While the pose encoder provides robust representations, its embeddings are not directly aligned with the semantic feature space of LLMs~\cite{chen2024motionllm,li2024llavaonevision,lin2024video}. To bridge this modality gap, we introduce a lightweight projection module that maps pose embeddings into the LLM’s embedding space. As shown in Fig.~\ref{fig:method}, this module is implemented as a two-layer multilayer perceptron (MLP), referred to as the \textit{pose translator}.  
The translated features are then passed to the multimodal LLM $\mathcal{Q}$, which also accepts optional video inputs $V$ through its visual encoder and text prompts $P$ through its language interface. This unified architecture allows the LLM to combine motion, video, and text information while maintaining modularity.
Unlike previous approaches~\cite{fang2025humocon,li2025human} that enforce tight alignment between motion and video modalities, we treat video as an auxiliary cue that enhances motion understanding, providing contextual grounding without constraining the motion representation space.

\subsection{Interpreting Fine-grained Timing in Motion}
\label{subsec:meth_temp}
The pose translator produces an embedding $m$ that is aligned with the token space of the LLM $\mathcal{Q}$. This embedding can be inserted into a textual query, for example:  
\textit{“In what position is this person \textless pose\textgreater?”},  
where \textless pose\textgreater{} is replaced by the pose embedding $m$. \(Q\) then processes the resulting prompt.

A naive extension from single-pose reasoning to motion reasoning would concatenate multiple pose embeddings, forming prompts such as  
\textit{“What is this person doing \textless pose\textgreater{} ... \textless pose\textgreater{}?”}. However, this concatenation introduces two key limitations.  
First, it neglects the precise timing of each pose, making it difficult for the model to infer when specific movements occur within the sequence.  
Second, the temporal resolution of such an encoding is fixed, as there is no mechanism to indicate variations in motion frame rate. Consequently, motions that are temporally downsampled to fit within the LLM’s context window, such as long sequences, are misinterpreted as faster or shorter actions.

To overcome these limitations, we introduce explicit timestamp conditioning. Each pose is concatenated with its corresponding timestamp, producing temporally grounded prompts such as:  
\textit{“What is this person doing? Time: 0.00\,s \textless pose\textgreater{}, Time: 0.03\,s \textless pose\textgreater{}, ..., Time: 5.03\,s \textless pose\textgreater{}.”}  
Here, the token “\textit{Time}” serves as a special indicator of pose timing, and the sampling interval can be flexibly adjusted. Timestamps denote elapsed seconds from the start of the input clip and retain their original values after temporal subsampling. This explicit timestamping provides the LLM with fine-grained temporal structure, supporting reasoning about motion order, duration, and rhythm.
Furthermore, it supports variable frame rates and adaptive temporal sampling strategies, allowing the model to process long motion sequences without losing temporal consistency. The shared temporal reference also facilitates reasoning over multi-person interactions by aligning poses from different individuals in time.  

\begin{figure*}[t!]
    \centering
    \includegraphics[width=\linewidth]{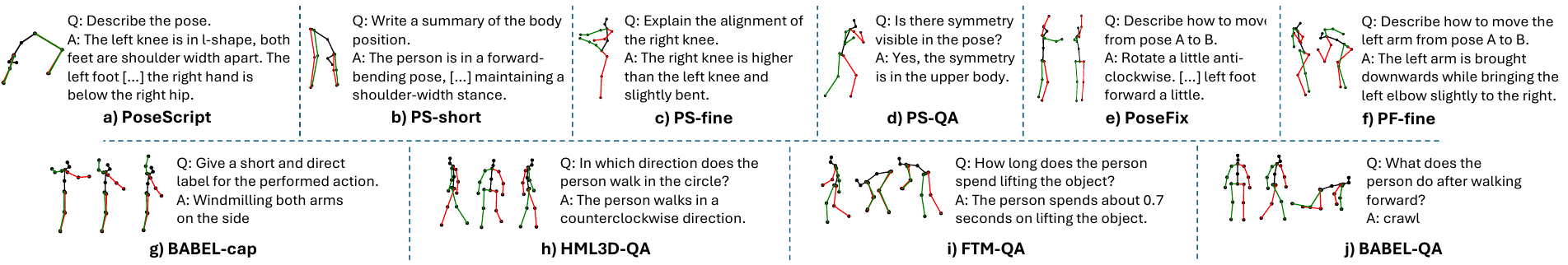}
    \caption{Training data examples. Pose-related datasets are shown in the top row and motion-related datasets in the bottom row. The modules provide complementary supervision, ranging from pose captions and pose QA to motion captions and temporally grounded motion QA. We analyze how these supervision components and training strategies affect skeletal motion-language reasoning.}
    \label{fig:training_data}
\end{figure*}

\subsection{Training with Pose-to-Motion Supervision}
We train the motion-language model using a diverse set of pose- and motion-level supervision modules derived from existing pose and motion datasets. These modules vary along two axes: the input type, ranging from isolated poses to full motion sequences, and the task type, ranging from captioning to question answering. This design lets us analyze how different supervision components contribute to skeletal motion-language reasoning, while also allowing the same data to be used either in a single training stage or in a staged schedule. Unlike recent motion-LLM methods~\cite{chen2024motionllm, fang2025humocon, li2025human}, which primarily adapt LLMs with motion-level QA supervision, our training mixture also includes pose-level captioning and QA. Our supervision modules are derived from standard sources already used in prior motion-LLM work: AMASS~\cite{mahmood2019amass}, BABEL~\cite{punnakkal2021babel}, and HumanML3D~\cite{humanml3d}. Thus, the gains do not stem from access to a larger motion corpus. Instead, we construct complementary pose- and motion-level annotations from the same underlying data, grounding both individual body configurations and their temporal relationships. Fig.~\ref{fig:training_data} shows representative examples for all modules.

\subsubsection{Pose Data Modules} We take samples from two human pose–text datasets: PoseScript~\cite{delmas2024posescript} and PoseFix~\cite{delmas2024posefixcorrecting3dhuman}. Both datasets use human poses from AMASS~\cite{mahmood2019amass} and provide rule-based machine generated textual descriptions. 
To further enrich the prompt diversity of these datasets and enhance the understanding of human pose data, we propose multiple data augmentations, resulting in various data modules that we list below.

We re-annotate PoseScript and PoseFix using an open-source LLM $\mathcal{F}$ to increase linguistic and prompt diversity. 
PoseScript provides multiple alternative descriptions for the same pose. For each pose, we provide two of these descriptions to $\mathcal{F}$ together with an instruction describing the desired output and hold out a third, independent PoseScript description for consistency filtering. Generated annotations are filtered with an LLM-as-a-judge consistency check against the held-out description, following prior re-annotation practices~\cite{wang2023self, zhang2025videollama3}. 
This procedure produces PS-short, PS-fine, and PF-fine; the supplement provides further details.

\textit{PS-short.}
To provide concise pose-level supervision, we condense PoseScript's multi-sentence descriptions into one-sentence summaries. This module produces short, semantically clear captions.

\textit{PS-fine and PF-fine.}
Inspired by image QA datasets~\cite{yuan2025ospreypixelunderstandingvisual,lin2025dra}, we generate fine-grained body-part QA pairs from PoseScript and PoseFix by instructing $\mathcal{F}$ to convert local limb and joint descriptions into questions and answers.
These new modules link language to detailed spatial pose configurations, improving both structural interpretability and instruction-following ability.

\textit{PS-QA.}
We also generate structured QA pairs directly from 3D pose geometry, covering body symmetry, joint distances, joint angles, and foot positioning. Because the answers are computed from pose coordinates, this module provides automatically verifiable supervision for spatial body reasoning and for linking pose cues to language. We balance referenced body parts and answer distributions to improve coverage and avoid bias. Further implementation details are provided in Sec.~\ref{sec:app_ps-qa}.

\subsubsection{Motion Data Modules and Temporal Alignment}
Following previous work~\cite{chen2024motionllm, fang2025humocon, li2025human}, we prompt an LLM to generate complex QA pairs for motions sampled from HumanML3D~\cite{humanml3d} (HML3D-QA) and include samples from BABEL-QA in our training data. As these QA sources provide limited coverage of simple motion descriptions and temporally detailed questions, we introduce BABEL-cap and FTM-QA to complement them.

\textit{BABEL-cap.}
To introduce simple motion-level supervision, we construct this data module from AMASS~\cite{mahmood2019amass} and BABEL~\cite{punnakkal2021babel}.  
Short clips extracted from AMASS are labeled with their corresponding BABEL actions to form concise question-answer pairs of the form \textit{``What action is being performed?''}. This module complements HML3D-QA by providing simple motion-level captioning.

\textit{FTM-QA.}
We introduce Fine-grained Temporal Motion QA (FTM-QA) to supervise questions about action duration, start and end times, and temporal ordering. The module follows the LLM-based QA generation procedure of~\cite{chen2024motionllm}, originally applied to HumanML3D, but adapts it to BABEL's temporally dense action annotations. This produces motion-text pairs that explicitly support temporal alignment and fine-grained motion reasoning.

\subsubsection{Training Strategies}
We consider two training strategies for the same supervision mixture. In the single-stage strategy, all data modules are combined and used throughout training. In the staged strategy, the modules are introduced according to input and task type, moving from pose-level supervision to motion-level supervision. In Stage~1, we use PoseScript and PS-short to establish pose-language alignment through direct pose-caption correspondences. Stage~2 adds PS-fine and PS-QA to increase linguistic diversity and fine-grained body-part reasoning. Stage~3 combines pose-reasoning modules with BABEL-cap, introducing simple motion-level supervision. Finally, Stage~4 adds the complex motion-level datasets HML3D-QA, BABEL-QA, and FTM-QA for temporally detailed motion reasoning over longer sequences. Additionally, a small subset of VCG-plus 112k~\cite{maaz2024videogptintegratingimagevideo} reinforces video-language grounding.

The single-stage and staged strategies use the same data modules. Therefore, comparing them allows us to distinguish the effect of supervision diversity from the effect of staging. Unless otherwise specified, we report the staged variant as \methodname because it performs best in our ablations.

\begin{table*}[tb]
\centering
\setlength{\tabcolsep}{11pt}
\begin{adjustbox}{width=\textwidth,center}
\begin{tabular}{@{}lccccccccc@{}}
\toprule
\multirow{2}{*}{Model} & \multirow{2}{*}{Overall} & \multicolumn{3}{c}{Query Type} & \multicolumn{4}{c}{Temporal Filter} \\ 
\cmidrule(l){3-5} \cmidrule(l){6-9} 
& & Action & Direction & Body Part & Before & After & In Between & Other \\ 
\midrule
NSPose \cite{endo2023motion} & 0.578 & 0.627 & 0.598 & 0.325 & 0.531 & 0.594 & 0.590 & 0.609 \\
IMoRe II \cite{hummanQA} & 0.640 & 0.695 & 0.679 & 0.358 & 0.600 & 0.649 & 0.675 & 0.663 \\
MotionLLM \cite{chen2024motionllm}                                                             & 0.436                    & 0.517          & 0.354          & 0.154          & 0.427          & 0.368          & -              & 0.529          \\
LLaMo \cite{li2025human}                                                                & 0.458                    & 0.525          & 0.398          & 0.224          & 0.443          & 0.392          & -              & 0.518          \\
HuMoCon \cite{fang2025humocon}                                                                & 0.711                    & {\ul 0.809}          & {\ul 0.697}    & {\ul 0.623}    & {\ul 0.707}          & 0.635          & -              & {\ul 0.797}          \\ \midrule
\textbf{\methodname (2D)}                                                             & {\ul 0.750}              & \textbf{0.816}    & 0.681          & 0.550 &  0.677    & {\ul 0.758}    & \textbf{ 0.781}    & \textbf{ 0.806}    \\
\textbf{\methodname (3D)}                                                            & \textbf{0.758}           & 0.791 & \textbf{0.722} & \textbf{0.658}          & \textbf{0.710} & \textbf{0.766} & {\ul 0.750} & {\ul 0.797} \\ \bottomrule
\end{tabular}
\end{adjustbox}
\caption{BABEL-QA accuracy. Query Type specifies whether a question targets an action, direction, or body part. Temporal Filter groups questions by whether the answer is sought before, after, between, or without referenced events. All prior methods use 3D motion. Our 3D model performs best overall, while our 2D model remains competitive despite using only reprojected 2D motion.
}
\label{tab:babel_qa}
\end{table*}

\section{Experiments}

\label{sec:experiments}
\textbf{Training.} The pose encoder is pre-trained following the MotionBERT~\cite{zhu2023motionbert} data pipeline, using 3D datasets Human3.6M~\cite{human3.6m} and AMASS~\cite{mahmood2019amass}, and 2D datasets PoseTrack~\cite{andriluka2018posetrack} and InstaVariety~\cite{kanazawa2019instavariety}.
During pre-training, 15\% of joints are randomly masked, and additive noise sampled from a mixture of Gaussian and uniform distributions~\cite{chang2020poselifter} is applied to enhance robustness.

We adopt VideoLLaMA3-7B~\cite{zhang2025videollama3} as the multimodal LLM $\mathcal{Q}$ and fine-tune the motion-language model using the pose- and motion-level supervision modules described in Sec.~\ref{sec:method}. All constructed supervision modules exclude source motions from benchmark validation and test sets, including derived poses and clips (Sec.~\ref{sec:motion_curriculum}). Unless otherwise specified, \methodname refers to the staged variant, which performs best in our ablations. The LLM $\mathcal{Q}$ is optimized using LoRA~\cite{lora}.

For 3D input, each pose is represented by joint coordinates $(x, y, z)$. For 2D input, each joint is represented as $(x, y, c)$, where $c$ denotes detection confidence and is set to 1 for re-projected 3D data.
Motion sequences are sampled at 30~fps, except for CompMo, which uses 20~fps, and are uniformly subsampled to a maximum of 800 frames.

\noindent\textbf{Evaluation and Metrics.} We report prediction accuracy on BABEL-QA and HuMMan-QA. For BABEL-QA, following~\cite{fang2025humocon, chen2024motionllm, liu2023visualinstructiontuning}, classifier-based methods are evaluated with direct accuracy, while generative methods are evaluated with an LLM-based answer-matching metric. This LLM-based evaluation is used only for BABEL-QA; HuMMan-QA accuracy is computed directly following~\cite{hummanQA}. In Sec.~\ref{sec:gpt_eval}, we additionally report strict string-matching accuracy, which does not rely on GPT-based evaluation. The main comparison remains unchanged: our 3D model achieves 0.728 under strict matching, exceeding HuMoCon’s reported GPT-evaluated score of 0.711. For CompMo~\cite{demo_3dv}, we report standard dense-captioning metrics and temporal localization metrics, including SODA, CIDEr, METEOR, tIoU, and F1. Additional experiments and qualitative assessments are provided in Sec.~\ref{sec:training}.

\subsection{Evaluation on BABEL-QA}
We evaluate our method on the BABEL-QA benchmark to assess fine-grained motion understanding. All baselines use ground-truth 3D motion as input. We report results for both 3D motion input and 2D motion input, where the latter is obtained by reprojecting the 3D skeletons into the image plane using a pinhole camera model (see Sec.~\ref{sec:app_pinhole}), denoted as ``Ours (3D)'' and ``Ours (2D)'', respectively.

As shown in Tab.~\ref{tab:babel_qa}, \methodname achieves the best overall performance. The gains are especially clear on temporally filtered questions such as ``After'' and ``In Between'', indicating that explicit timestamp conditioning helps the model reason about action order. Compared with prior motion-LLM methods that rely on compressed motion tokens or key poses, our timestamped pose representation preserves the temporal origin of each pose, which is important for fine-grained motion QA.

Remarkably, even with the less informative 2D motion input, our method outperforms all previous 3D-based approaches and achieves competitive performance with our 3D variant. Since both ``Ours (2D)'' and ``Ours (3D)'' are fine-tuned from the same \methodname, this result underscores the strength of the unified pose encoder and training scheme, which generalize effectively across motion representations.

Beyond reprojected 2D poses, Sec.~\ref{sec:ntu} evaluates \methodname using 2D skeletons detected from real videos on NTU-RGB+D 120. The appendix further evaluates detected poses on an exercise-feedback task (Sec.~\ref{sec:app_downstream}). Together, these experiments support the applicability of \methodname to practical pose detections beyond the reprojected BABEL-QA setting.

\begin{table*}[bt]
\resizebox{\linewidth}{!}{%
\begin{tabular}{@{}lcccccccccccc@{}}
\toprule
\multirow{2}{*}{Model} & \multirow{2}{*}{Overall} & \multicolumn{4}{c}{Query Action} & \multicolumn{3}{c}{Query Direction} & \multicolumn{4}{c}{Query Body Part} \\ \cmidrule(l){3-6} \cmidrule(l){7-9} \cmidrule(l){10-13} 
\multicolumn{1}{c}{} &  & All & Before & After & BTW & All & Before & After & All & Before & After & BTW \\ \midrule
MotionCLIP-RNN \cite{hummanQA} & 0.623 & 0.601 & 0.559 & 0.541 & 0.583 & 0.750 & 1.000 & 0.500 & 0.667 & 0.638 & 0.616 & 0.694 \\
NSPose \cite{endo2023motion} & 0.691 & 0.700 & 0.686 & 0.610 & 0.729 & 0.822 & 0.425 & 0.833 & 0.677 & 0.620 & 0.639 & 0.833 \\
IMoRe I \cite{hummanQA} & 0.719 & 0.744 & 0.652 & 0.734 & {\ul 0.854} & 1.000 & 1.000 & 1.000 & 0.665 & 0.609 & 0.647 & {\ul 0.889} \\
IMoRe II \cite{hummanQA} & 0.730 & 0.746 & 0.648 & 0.739 & 0.813 & 1.000 & 1.000 & 1.000 & 0.717 & 0.636 & 0.703 & 0.861 \\ \midrule
\textbf{FiGMo (2D)} & {\ul 0.792} & {\ul 0.808} & {\ul 0.716} & {\ul 0.795} & 0.800 & 1.000 & 1.000 & 1.000 & {\ul 0.753} & {\ul 0.652} & {\ul 0.767} & \textbf{1.000} \\
\textbf{FiGMo (3D)} & \textbf{0.824} & \textbf{0.841} & \textbf{0.770} & \textbf{0.828} & \textbf{0.900} & 1.000 & 1.000 & 1.000 & \textbf{0.782} & \textbf{0.696} & \textbf{0.781} & \textbf{1.000} \\ \bottomrule
\end{tabular}%
}
\caption{HuMMan-QA accuracy. Columns group questions by the queried attribute---action, direction, or body part---and by temporal relation: All, Before, After, and Between (BTW). All methods are fine-tuned and evaluated following the HuMMan-QA protocol. Our method outperforms prior work with both 2D and 3D motion input.}
\label{tab:humman-qa}
\end{table*}  

\subsection{Evaluation on HuMMan-QA}
To evaluate generalization beyond BABEL, we follow prior work \cite{hummanQA} on the HuMMan-QA benchmark, which is built on the HuMMan-MoGen motion dataset \cite{huMMan-GEN}. We fine-tune our model on the HuMMan-QA training split following their proposed protocol \cite{hummanQA}. We compare with strong classification-based baselines, NSPose \cite{endo2023motion}, and IMoRe \cite{hummanQA}. The generative motion-language models do not report on HuMMan-QA and lack public training code, and are therefore excluded.

As shown in Tab.~\ref{tab:humman-qa}, our method achieves the best performance across all evaluation metrics. These results indicate that the timestamped pose representation and pose- and motion-level supervision transfer beyond BABEL-QA, demonstrating transfer to a different motion distribution.

\subsection{Evaluation on Dense Motion Captioning}
To directly evaluate temporal grounding and whether explicit timestamps preserve temporal scale, we test \methodname on the CompMo dense motion captioning benchmark~\cite{demo_3dv}. Unlike BABEL-QA and HuMMan-QA, which evaluate motion understanding through question answering, CompMo requires the model to both localize motion segments in time and generate a caption for each segment.

This setting is particularly informative for our representation. Ordinary concatenation of pose embeddings preserves their order but does not specify the time interval between consecutive poses. Consequently, changing the sampling rate changes the implied speed and duration of the motion. In contrast, \methodname associates every pose with its physical timestamp, preserving the temporal scale when the sequence is subsampled. CompMo therefore allows us to evaluate both temporal localization under a fixed sampling rate and robustness when the inference frame rate changes.

As shown in Tab.~\ref{tab:compmo}, both \methodname and its non-timestamped variant outperform UniMotion~\cite{unimotion} and DEMO~\cite{demo_3dv} at 20 fps, showing that the overall architecture and supervision already provide a strong basis for dense motion captioning. Timestamp grounding provides substantial additional gains, raising tIoU from 86.81 to 98.44 and F1 from 78.59 to 97.48 while improving every captioning metric. The largest gains occur in temporal localization, consistent with explicitly associating each pose with its temporal origin. This advantage is also visible in Sec.~\ref{sec:app_compmo_qual}, where, on examples presented in the original DEMO paper, \methodname recovers missed segments, better aligns temporal boundaries, and more precisely captures attributes such as speed, direction, and action style.

Both variants are trained at 20~fps and evaluated at the same rate for the standard comparison; the subsampling analysis changes only the inference sampling rate. Reducing the rate from 20 to 10~fps leaves the timestamped model nearly unchanged (97.48 to 96.77 F1), whereas the non-timestamped variant falls from 78.59 to 28.99 because it cannot represent the physical interval between poses. Notably, \methodname at 10~fps outperforms the non-timestamped model at 20~fps across every metric. Additional results at 5~fps are provided in Sec.~\ref{sec:app_compmo_fps}.

\begin{table*}[tb]
\centering
\begin{adjustbox}{width=\textwidth,center}
\begin{tabular}{@{}lcccccccccc@{}}
\toprule
\multirow{2}{*}{Model}
& \multirow{2}{*}{FPS}
& \multicolumn{7}{c}{Dense Captioning $\uparrow$}
& \multicolumn{2}{c}{Temporal $\uparrow$} \\
\cmidrule(lr){3-9}
\cmidrule(lr){10-11}
& & SODA & SODA-B & CIDEr & METEOR & ROUGE-L & BLEU@1 & BLEU@4
& tIoU & F1 \\
\midrule
UniMotion~\cite{unimotion}
& 20 & 0.61 & 12.81 & 1.01 & 0.43 & 0.85 & 0.78 & 0.00 & 36.14 & 4.00 \\
DEMO~\cite{demo_3dv}
& 20 & 17.85 & 64.40 & 134.44 & 16.41 & 24.05 & 23.90 & 11.00 & 77.94 & 58.21 \\
\midrule
\methodname w/o TS Ground
& 10 & 22.32 & 49.36 & 158.03 & 17.67 & 21.71 & 21.75 & 15.51 & 60.91 & 28.99 \\
\methodname w/o TS Ground
& 20 & 46.59 & 79.95 & 423.68 & 37.94 & 53.63 & 51.98 & 41.91 & 86.81 & 78.59 \\
\midrule
\textbf{\methodname}
& 10 & \textbf{54.56} & 90.75 & \textbf{508.79} & \textbf{41.76}
& \textbf{63.79} & \textbf{61.42} & \textbf{49.83} & 97.78 & 96.77 \\
\textbf{\methodname}
& 20 & 50.98 & \textbf{91.00} & 456.95 & 38.77
& 60.48 & 58.41 & 45.38 & \textbf{98.44} & \textbf{97.48} \\
\bottomrule
\end{tabular}
\end{adjustbox}
\caption{Comparison on the CompMo dense motion captioning benchmark~\cite{demo_3dv}, all in 3D. FPS denotes the inference sampling rate. In addition to comparing with prior work at 20 fps, we evaluate matched variants with and without timestamp grounding at 10 and 20 fps.}
\label{tab:compmo}
\end{table*}

\begin{table*}[tb]
\centering
\setlength{\tabcolsep}{11pt}
\begin{adjustbox}{width=\textwidth,center}
\begin{tabular}{@{}lcccccccc@{}}
\toprule
\multirow{2}{*}{Model} & \multirow{2}{*}{Overall} & \multicolumn{3}{c}{Query Type} & \multicolumn{4}{c}{Temporal Filter} \\ 
\cmidrule(l){3-5} \cmidrule(l){6-9} 
& & Action & Direction & Body Part & Before & After & In Between & Other \\ 
\midrule
w/o TS Ground & 0.711 & 0.757 & 0.625 & 0.617 & 0.621 & 0.702 & 0.734 & 0.797 \\ 
BABEL-QA only & 0.443 & 0.481 & 0.389 & 0.342 & 0.383 & 0.435 & 0.594 & 0.500 \\
w/o Re-annotated Sup. & 0.642 & 0.695 & 0.556 & 0.517 & 0.633 & 0.629 & 0.625 & 0.662 \\
w/o Pose Sup. & 0.720 & 0.764 & 0.667 & 0.592 & 0.673 & 0.746 & 0.734 & 0.738 \\  
w/o Motion Sup. & 0.724 & 0.797 & 0.597 & 0.558 & 0.665 & 0.742 & 0.781 & 0.759 \\
Single-stage & 0.751 & 0.793 & 0.708 & 0.617 & 0.698 & 0.758 & 0.828 & 0.790 \\
Complete staged model & 0.758 & 0.791 & 0.722 & 0.658 & 0.710 & 0.766 & 0.750 & 0.797 \\ 
\bottomrule
\end{tabular}
\end{adjustbox}
\caption{Ablations on BABEL-QA with 3D motion input. Timestamp grounding improves temporal reasoning; re-annotated and diverse pose- and motion-level supervision drive performance, while staged training provides a smaller gain.}
\label{tab:ablations}
\end{table*}

\subsection{Evaluation with Detected 2D Poses}
\label{sec:ntu}
To evaluate performance beyond clean or reprojected skeletons, we test
\methodname on NTU-RGB+D 120 using 2D poses detected directly from real
videos. We follow the established one-shot action-recognition protocol~\cite{zhu2023motionbert, liu2019ntu}.
Among methods evaluated using detected 2D skeletons, MotionBERT~\cite{zhu2023motionbert} achieves
67.4\% accuracy, while \methodname using detected skeletal motion reaches
69.2\%.

We additionally evaluate the optional video pathway. Video-only input
achieves 66.8\%, whereas combining detected motion and video increases
accuracy to 77.4\%. These results show that skeletal motion is informative
independently and that RGB provides complementary contextual information.
The complete comparison and evaluation protocol are provided in Sec.~\ref{sec:app_ntu}.

\subsection{Component Ablations}
We conduct ablations on BABEL-QA to analyze timestamp grounding, pose-level supervision, motion-level supervision, and the effect of staging the same supervision modules. All variants use 3D motion input and are fine-tuned on BABEL-QA. The results are summarized in Tab.~\ref{tab:ablations}.

Removing timestamp grounding decreases overall accuracy from 0.758 to 0.711. The degradation is most visible for temporally sensitive query types such as ``Before'', ``After'', and ``In Between'', while queries without an explicit temporal filter (``Other'') remain relatively stable.

Removing all of our re-annotated supervision modules reduces overall accuracy from 0.758 to 0.642. In this variant, the model is trained exclusively on existing datasets, without any supervision modules constructed through our re-annotation procedures. The substantial decrease confirms that the proposed modules contribute useful supervision rather than redundant annotations.

Removing pose-level or additional motion-level supervision also hurts performance, reaching 0.720 and 0.724, respectively. The pose ablation is particularly informative: although BABEL-QA evaluates motion QA, removing static pose supervision substantially reduces performance, suggesting that pose-level supervision is important for downstream motion understanding. The motion ablation further shows that BABEL-QA fine-tuning captures part of the required reasoning, but broader motion supervision from our motion modules improves detailed understanding. When removing both pose- and motion-level supervision using only BABEL-QA fine-tuning, performance drops to 0.443, further confirming that the broader supervision mixture is essential.

Finally, to isolate the effect of staging from the effect of data composition, we train a single-stage variant using the same supervision modules as the complete model. 
This variant reaches 0.751 overall accuracy, close to the 0.758 obtained by the staged model. Thus, while the staged schedule performs best, the small gap indicates that the main performance driver is the diverse pose- and motion-level supervision rather than the staging itself.

\section{Conclusion}
We presented \methodname, an LLM-based model that represents human motion as timestamp-grounded skeletal pose sequences. Explicit pose-time alignment preserves temporal scale under subsampling and supports reasoning over action order, duration, and temporally localized events. To ground individual body configurations and their changes over time, we constructed pose-to-motion supervision spanning pose- and motion-level language tasks. Our ablations show that this supervision drives the main gains, while staging provides a smaller additional benefit. Across motion question answering, dense motion captioning, and action recognition with detected poses, \methodname achieves state-of-the-art performance using 2D and 3D skeletal inputs.

Future work could enrich timestamped motion-language reasoning with scene and object context. Skeletal poses capture body dynamics compactly, but omit information about objects, contact, and 3D scene layout that may be important for interaction understanding. Adaptive temporal sampling is also a promising direction for scaling to long sequences while preserving short but important actions.

{
    \small
    \bibliographystyle{ieeenat_fullname}
    \bibliography{main}
}
\clearpage
\appendix
Sec.~\ref{sec:training} expands on the experiments and evaluation protocols from the main paper. Sec.~\ref{sec:data_generation} details the construction of the pose- and motion-level supervision modules. Finally, Secs.~\ref{sec:limitations_impact}--\ref{sec:assets} discuss limitations, broader impact, compute resources, and assets used in this work.

\section{Extra Experiments, Training \& Evaluation Details}
\label{sec:training}

This section provides additional experiments that complement the main paper, the evaluation details are described in this section but not listed here:

\begin{itemize}\itemsep2pt
    \item \textbf{Qualitative Comparison on Dense Motion Captioning} (Sec.~\ref{sec:app_compmo_qual}). We qualitatively compare results with DEMO~\cite{demo_3dv}, showing that \methodname performs better at both temporal localization and captioning.

    \item \textbf{CompMo frame-rate analysis} (Sec.~\ref{sec:app_compmo_fps}).
    We evaluate timestamped and non-timestamped variants under temporal subsampling.

    \item \textbf{NTU-RGB+D 120 comparison and protocol}
    (Sec.~\ref{sec:app_ntu}).
    We provide the complete comparison and one-shot evaluation details for the
    detected-pose experiment summarized in Sec.~\ref{sec:ntu}.

    \item \textbf{Downstream exercise feedback} (Sec.~\ref{sec:app_downstream}). 
    We test whether explicit skeletal motion representations help provide feedback on exercise execution from detected 2D poses.

    \item \textbf{Qualitative examples} (Sec.~\ref{sec:app_qualitative}). 
    We show examples of timestamp-aware motion reasoning, action decomposition, and semantic pose interpretation.

    \item \textbf{BABEL-QA GPT-evaluation analysis} (Sec.~\ref{sec:gpt_eval}). 
    We compare strict string matching with GPT-based semantic matching to verify that the LLM-based BABEL-QA evaluator reflects meaningful answer correctness.


\end{itemize}

\subsection{Dense Motion Captioning (CompMo) Qualitative Comparison}
\label{sec:app_compmo_qual}
\begin{figure*}[h]
    \centering

    \begin{subfigure}{\linewidth}
        \centering
        \includegraphics[width=\linewidth]{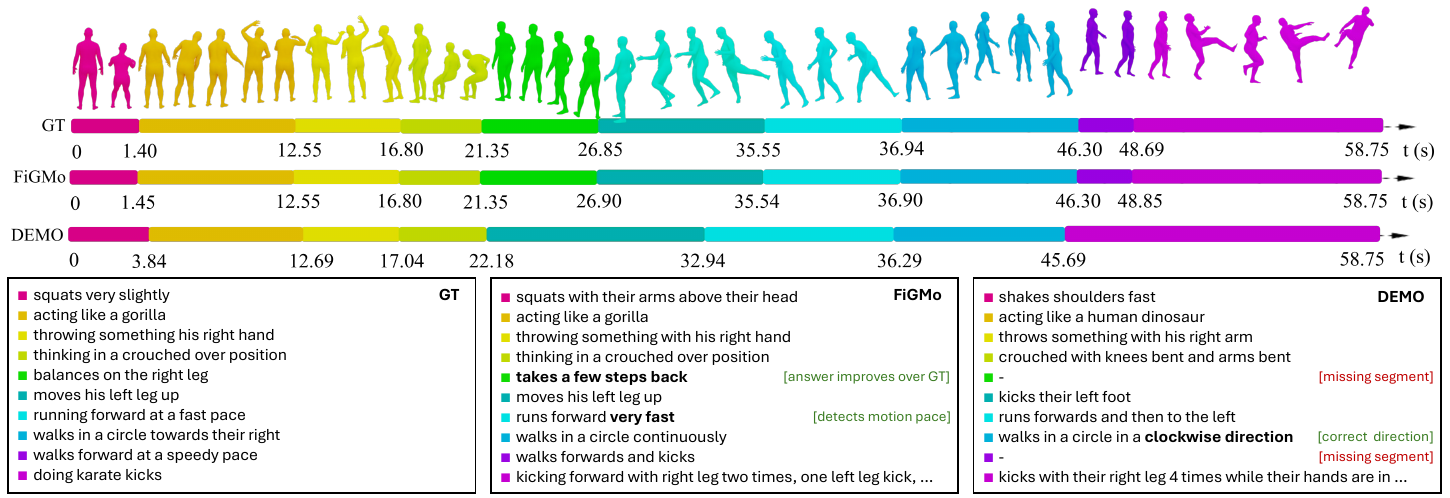}
        \caption{}
        \label{fig:compmo_vis1}
    \end{subfigure}

    \vspace{0.5em}

    \begin{subfigure}{\linewidth}
        \centering
        \includegraphics[width=\linewidth]{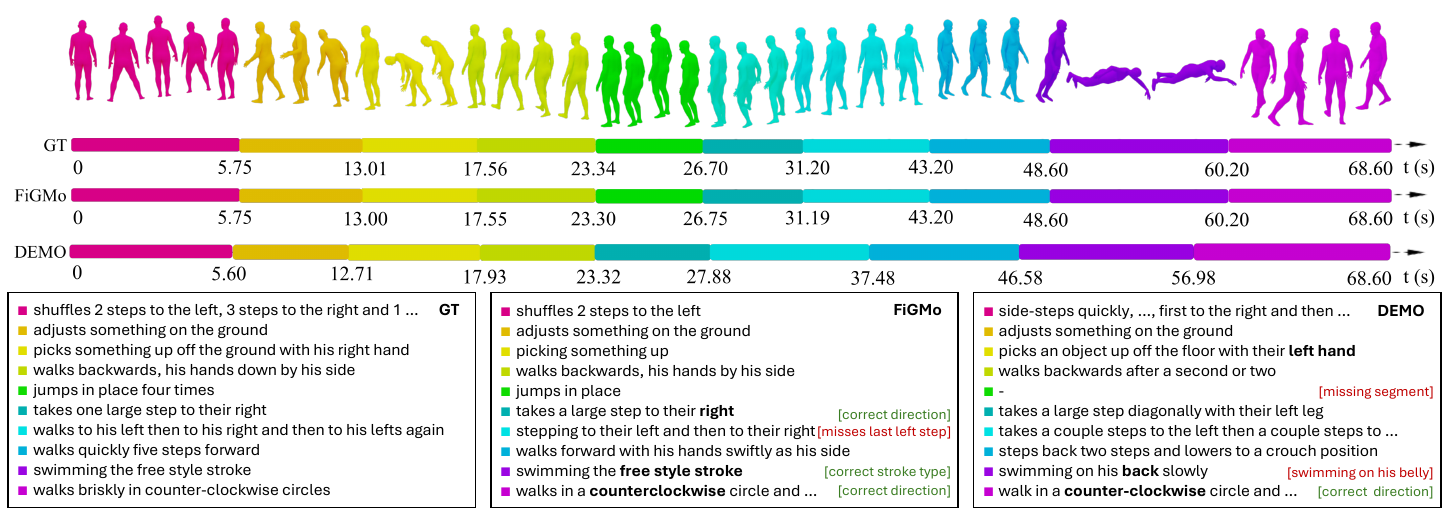}
        \caption{}
        \label{fig:compmo_vis2}
    \end{subfigure}

    \caption{Qualitative comparison of \methodname with DEMO~\cite{demo_3dv} on the CompMo dense motion captioning dataset~\cite{demo_3dv}. 
    We use the original qualitative examples and pose renderings from the DEMO paper to avoid cherry-picking examples in favor of \methodname. 
    Across these examples, \methodname shows stronger temporal localization and more precise captions.}
    \label{fig:compmo_vis}
\end{figure*}

Fig.~\ref{fig:compmo_vis} shows a qualitative comparison between \methodname and DEMO~\cite{demo_3dv} on CompMo~\cite{demo_3dv}. 
To avoid cherry-picking examples in favor of \methodname, we use the same qualitative examples selected in the original DEMO paper. 
The pose renderings are also reused from the original DEMO visualization; we only add the predictions of \methodname for comparison.

Consistent with the quantitative results in Tab.~\ref{tab:compmo}, \methodname shows stronger temporal localization than DEMO. 
In these examples, \methodname detects all annotated segments, while DEMO misses some segments, and the predicted temporal boundaries are generally better aligned with the ground truth. 
For captioning, DEMO already produces reasonable descriptions, but \methodname captures several motion concepts more precisely. 
For example, in Fig.~\ref{fig:compmo_vis1}, \methodname captions \textcolor{green}{\rule{0.65em}{0.65em}} as ``\textit{takes a few steps back}'', which appears more specific to the visualized motion than the ground-truth caption. 
Similarly, \textcolor{SkyBlue}{\rule{0.65em}{0.65em}} is captioned as ``\textit{runs forwards very fast}'', reflecting the fast pace described in the ground truth. 
Fig.~\ref{fig:compmo_vis2} shows a similar pattern: for \textcolor{Plum}{\rule{0.65em}{0.65em}}, \methodname correctly identifies the swimming stroke as ``\textit{freestyle}'', whereas DEMO does not.

\subsection{CompMo Frame-Rate Analysis}
\label{sec:app_compmo_fps}

We evaluate \methodname and its non-timestamped variant at inference sampling
rates of 5, 10, and 20 fps. Both variants use the same architecture,
supervision, and training procedure; only timestamp grounding and the inference
sampling rate differ.

As shown in Tab.~\ref{tab:compmo_fps}, the timestamped model remains robust
under increasingly aggressive subsampling. At 5 fps, it retains 90.36 tIoU and
83.96 F1, whereas the non-timestamped variant falls to 53.33 tIoU and 15.90
F1. This supports the interpretation that explicit timestamps preserve the
physical temporal scale when the interval between sampled poses changes.

\begin{table*}[h]
\centering
\begin{adjustbox}{width=\textwidth,center}
\begin{tabular}{@{}lcccccccccc@{}}
\toprule
\multirow{2}{*}{Model}
& \multirow{2}{*}{FPS}
& \multicolumn{7}{c}{Dense Captioning $\uparrow$}
& \multicolumn{2}{c}{Temporal $\uparrow$} \\
\cmidrule(lr){3-9}
\cmidrule(lr){10-11}
& & SODA & SODA-B & CIDEr & METEOR & ROUGE-L & BLEU@1 & BLEU@4
& tIoU & F1 \\
\midrule
\methodname w/o TS Ground
& 5 & 6.16 & 35.33 & 29.87 & 5.07 & 6.58 & 6.84 & 2.22 & 53.33 & 15.90 \\
\methodname w/o TS Ground
& 10 & 22.32 & 49.36 & 158.03 & 17.67 & 21.71 & 21.75 & 15.51 & 60.91 & 28.99 \\
\methodname w/o TS Ground
& 20 & 46.59 & 79.95 & 423.68 & 37.94 & 53.63 & 51.98 & 41.91 & 86.81 & 78.59 \\
\midrule
\textbf{\methodname}
& 5 & 30.64 & 79.06 & 266.08 & 25.30 & 40.41 & 39.43 & 24.77 & 90.36 & 83.96 \\
\textbf{\methodname}
& 10 & \textbf{54.56} & 90.75 & \textbf{508.79} & \textbf{41.76}
& \textbf{63.79} & \textbf{61.42} & \textbf{49.83} & 97.78 & 96.77 \\
\textbf{\methodname}
& 20 & 50.98 & \textbf{91.00} & 456.95 & 38.77
& 60.48 & 58.41 & 45.38 & \textbf{98.44} & \textbf{97.48} \\
\bottomrule
\end{tabular}
\end{adjustbox}
\caption{CompMo performance under inference-time temporal subsampling. Explicit timestamp grounding substantially improves robustness when the sampling rate differs from the 20-fps training setting.}
\label{tab:compmo_fps}
\end{table*}

\subsection{Evaluation on NTU-RGB+D 120}
\label{sec:app_ntu}
We further evaluate our model on the NTU-RGB+D 120 dataset under the one-shot recognition protocol~\cite{zhu2023motionbert, liu2019ntu}. This one-shot setting is challenging, as i) the model must generalize to unseen actions from a single labeled instance, and ii) it also contains various two-person actions.
Following prior work, we use 2D skeleton detections as motion input and additionally evaluate a multimodal variant that combines motion and video. The models are fine-tuned on an auxiliary set of 100 classes and one example per class from the 20 test categories. We report classification accuracy over the 20 classes in the one-shot evaluation set.

As shown in Tab.~\ref{tab:ntu120-1shot}, our model achieves 69.2\% accuracy using only motion input, surpassing the compared methods.
When video is added as an auxiliary modality, performance rises to 77.4\%, achieving the highest accuracy among the compared methods.
To our knowledge, this is the first evaluation of combined motion-video modeling in the one-shot setting of~\cite{liu2019ntu}, suggesting that visual context can effectively complement motion features for action understanding.

To examine the role of video in the multimodal model, we also fine-tune and evaluate \methodname using video input alone.
This variant achieves 66.8\% accuracy, which is lower than the motion-only model. This suggests that explicitly representing pose trajectories is beneficial compared with relying on sparsely sampled RGB frames alone.

Finally, since NTU includes two-person actions, this experiment also demonstrates that the model can process multi-person motion sequences in interaction-based activities.

\begin{table}[h]
\centering
\begin{adjustbox}{max width=\linewidth,center}
\begin{tabular}{@{}lc@{}}
\toprule
Model & Acc. \\ 
\midrule
ST-LSTM + AvgPool~\cite{liu2016spatio} & 42.9 \\
APSR~\cite{liu2019ntu} & 45.3 \\
TCN Oneshot~\cite{sabater2021one} & 46.5 \\
SL-DML~\cite{memmesheimer2021sl} & 50.9 \\
Skeleton-DML~\cite{memmesheimer2022skeleton} & 54.2 \\
MotionBERT~\cite{zhu2023motionbert} & 67.4 \\
\midrule
\textbf{\methodname (2D Motion)} & 69.2 \\ 
\textbf{\methodname (2D Motion + Video)} & \textbf{77.4} \\ 
\bottomrule
\end{tabular}
\end{adjustbox}
\caption{Comparison on the one-shot NTU-RGB+D 120 dataset using 2D skeleton detections.}
\label{tab:ntu120-1shot}
\end{table}

\subsection{Evaluation on Downstream Task}
\label{sec:app_downstream}
As an additional downstream evaluation, we test \methodname on an exercise-feedback task derived from the QEVD-Coach dataset~\cite{panchal2024saysayitlive}. This dataset contains videos of people performing physical exercises together with question-answer pairs describing exercise execution. For the downstream application test, we report language similarity metrics METEOR~\cite{banerjee-lavie-2005-meteor}, ROUGE-L~\cite{lin-2004-rouge}, and BERTScore~\cite{zhang2020bertscoreevaluatingtextgeneration}.

From the original 300k samples, we select a subset of 30k examples and construct a train-test split. For the motion-based setting, 2D skeletons are extracted from the videos using MMPose \cite{contributors2020openmmlab}. All models are fine-tuned on the same data yet operate on different input modalities. The video and text-only training settings follow the protocols proposed in VideoLLaMA3~\cite{zhang2025videollama3} and Qwen2.5~\cite{qwen2025qwen25technicalreport}. Note all compared methods use Qwen2.5 as the backbone model, ensuring that performance differences primarily stem from the input modality.

Tab.~\ref{tab:qevd} shows the results. Our motion-based model outperforms VideoLLaMA3, which operates directly on raw video frames. The results suggest that explicit skeletal motion representations provide strong signals for evaluating exercise execution. Qualitatively, the video model struggles with subtle differences between visually similar exercises (e.g., planks versus push-ups), which are difficult to infer from temporally sparse sampled video frames. The text-only model operates without video or motion input and therefore provides a lower bound.

\begin{table}[h]
\centering
\begin{adjustbox}{width=\linewidth,center}
\begin{tabular}{@{}lcccc@{}}
\toprule
Model & Modality & METEOR & ROUGE-L & BERTScore \\
\midrule
Qwen2.5~\cite{qwen2025qwen25technicalreport} 
    & Text 
    & 0.859 
    & 0.849 
    & 0.978 \\
VideoLLaMA3~\cite{zhang2025videollama3} 
    & Video 
    & 0.941 
    & 0.935 
    & 0.990 \\
\textbf{\methodname} 
    & 2D Motion 
    & \textbf{0.946} 
    & \textbf{0.940} 
    & \textbf{0.991} \\
\bottomrule
\end{tabular}
\end{adjustbox}
\caption{Comparison on the downstream exercise-feedback task on QEVD-Coach. We compare \methodname using detected 2D skeletal motion against a video LLM operating on raw video, and include a text-only LLM as a lower-bound baseline.}
\label{tab:qevd}
\end{table}

\subsection{Qualitative Results}
\label{sec:app_qualitative}
\begin{figure*}[h]
    \centering
    \includegraphics[width=1.0\linewidth]{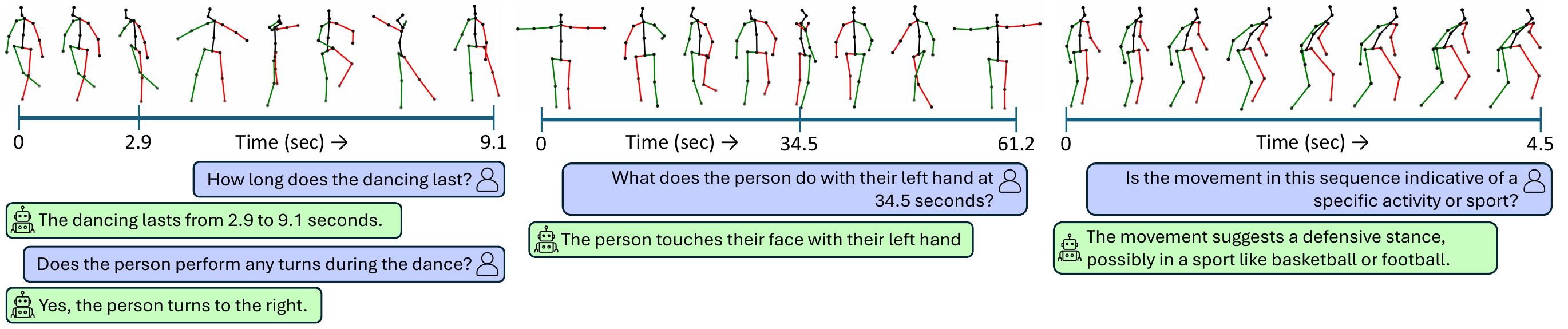}
    \caption{Qualitative examples showing fine-grained temporal motion reasoning (left), action decomposition from a minute-long complex motion sequence (middle), and semantic understanding of human activity (right).}
    \label{fig:qualitative}
\end{figure*}

We present qualitative examples from our method to illustrate its capability in fine-grained temporal and semantic motion understanding.  
Fig.~\ref{fig:qualitative} shows representative cases of instructional queries on test samples.
The left example demonstrates that the method not only identifies actions but also understands their temporal structure.  
For a dancing clip, the method correctly decomposes a complex movement into smaller sub-actions when prompted, showing that large composite motions can be parsed into meaningful fine-grained segments.  
The middle example shows that given a 60-second motion clip, the method accurately links a specific timestamp to a momentary action lasting less than a second.  
The right example illustrates that the method can associate the pose with relevant sports contexts (e.g., basketball or football), demonstrating its ability to ground pose semantics in broader human activity domains.
Together, these qualitative results illustrate the strengths of our method in timestamp-aware motion reasoning, action decomposition, and semantic pose interpretation.

\subsection{BABEL-QA: Effect of GPT-3.5-turbo Evaluation}

\label{sec:gpt_eval}
We compare strict string matching with a semantics-aware GPT-based evaluation (GPT-eval), which uses GPT-3.5-turbo to judge semantic equivalence between the model output and the ground truth. The evaluation follows the protocol used in prior work on BABEL-QA; further details are provided in Sec.~\ref{subsubsec:gpt-eval}. Tab.~\ref{tab:gpt-eval} reports the results.

\textit{Semantic Equivalence Gains.}
GPT-eval consistently yields higher scores because it recognizes semantic similarity (e.g., \emph{``jog''} vs.\ \emph{``run''}). Our 2D model improves by 4.3 percentage points overall (0.707~$\rightarrow$~0.750), especially in categories such as Action and Body Part, where minor lexical variations are common.

\textit{Model Superiority.}
Even under strict string matching---a harsher metric than used for the HuMoCon baseline---our 3D model achieves an Overall Score of 0.728, exceeding HuMoCon’s GPT-evaluated score of 0.711. Under GPT-eval, our model reaches 0.758, establishing clear superiority across evaluation protocols.

\textit{Evaluation Robustness.}
To verify that GPT-eval does not inflate results indiscriminately, we examine the Direction category, which has a closed set of four mutually exclusive labels (right, left, forward, backward). Scores are identical under both metrics (2D: 0.681, 3D: 0.722), confirming that GPT-eval only relaxes evaluation when semantic interpretation is genuinely required.

\begin{table*}[h]
\centering
\setlength{\tabcolsep}{11pt}
\caption{Comparison of strict string matching and GPT-3.5-turbo evaluation on BABEL-QA. HuMoCon uses GPT-eval. Our models outperform baselines under both evaluation schemes.}
\label{tab:gpt-eval}
\resizebox{\textwidth}{!}{%
\begin{tabular}{@{}lccccccccc@{}}
\toprule
\multirow{2}{*}{Model} & \multirow{2}{*}{Eval} & \multirow{2}{*}{Overall} & \multicolumn{3}{c}{Query Type} & \multicolumn{4}{c}{Temporal Filter} \\
\cmidrule(l){4-6} \cmidrule(l){7-10}
 &  &  & Action & Direction & Body Part & Before & After & In Between & Other \\ \midrule
HuMoCon (3D) & GPT-eval & 0.711 & 0.809 & 0.697 & 0.623 & 0.707 & 0.635 & - & 0.797 \\ \midrule
\methodname (2D) & Str.-match & 0.707 & 0.785 & 0.681 & 0.400 & 0.637 & 0.718 & 0.750 & 0.759 \\
\methodname (2D) & GPT-eval &  0.750 & 0.816    & 0.681          & 0.550 &  0.677    & 0.758    &  0.781    &  0.806 \\ \midrule
\methodname (3D) & Str.-match & 0.728 & 0.770 & 0.722 & 0.550 & 0.677 & 0.742 & 0.719 & 0.759 \\
\methodname (3D) & GPT-eval & 0.758   & 0.791 & 0.722 & 0.658          & 0.710 & 0.766 & 0.750 & 0.797 \\ \bottomrule
\end{tabular}%
}
\end{table*}



\subsection{Evaluation on ActivityNet-QA}
\label{sec:app_activitynet}
For completeness, we evaluate our model on ActivityNet-QA, as prior motion-oriented works such as MotionLLM~\cite{chen2024motionllm} and HuMoCon~\cite{fang2025humocon} report results on this benchmark using video-only input. ActivityNet-QA is a general-purpose video question answering dataset and is not specifically designed to assess detailed human motion reasoning. Only a small fraction of the questions (approximately 10\%) are directly related to human motion, and an even smaller portion requires the fine-grained temporal understanding targeted by our approach.

On ActivityNet-QA, we follow~\cite{chen2024motionllm} and use LLM-based evaluation. The results in Tab.~\ref{tab:activitynet} show that our method achieves comparable or slightly better performance than previous motion-oriented works, although our model is primarily designed for motion-based reasoning and the video is only an auxiliary modality.

\begin{table}[h]
\centering
\setlength{\tabcolsep}{18pt}
\begin{adjustbox}{width=\linewidth,center}
\begin{tabular}{lcc}
\toprule
Model & Acc. (\%) $\uparrow$ & Score $\uparrow$ \\ 
\midrule
MotionLLM~\cite{chen2024motionllm} & 53.3 & 3.5 \\
HuMoCon~\cite{fang2025humocon} & 54.2 & \textbf{3.6} \\
\textbf{\methodname} & \textbf{54.4} & \textbf{3.6} \\ 
\bottomrule
\end{tabular}
\end{adjustbox}
\caption{Comparison on the ActivityNet-QA benchmark against prior motion-understanding LLMs. Although video is used only as an auxiliary input modality in our model, we achieve on-par or slightly better performance compared to methods designed for video-only inputs.}
\label{tab:activitynet}
\end{table}

\subsection{GPT Evaluation Prompts}
\label{subsubsec:gpt-eval}

\paragraph{BABEL-QA Evaluation.} To evaluate generative answers on BABEL-QA, we follow \cite{chen2024motionllm, fang2025humocon, li2025human, li2025chatmotion} to use a semantics-aware GPT-based evaluation protocol. For each question–answer pair, we query a GPT model and ask it to judge whether the model's prediction is semantically consistent with the ground-truth answer. This allows synonyms, paraphrases, and linguistically natural variations to be considered correct when appropriate.

All evaluations use \texttt{gpt-3.5-turbo-0125}, following previous work. Queries are executed via the batch-completions API. For each evaluation sample, we prepare a pair of prompts:  
(i) a system prompt describing the evaluation rules, and  
(ii) a user prompt containing the detailed instruction, question, predicted answer, and ground truth. The GPT API returns one semantic similarity score per item, which we parse and aggregate into the final accuracy.

The GPT model is instructed to output only a Python dictionary string of the form:  
\texttt{\{'score': <float>\}},  
where the score lies between 0 and 1. This strict output structure simplifies automatic parsing and metric computation. We show the full prompt template used for GPT-based scoring in Fig.~\ref{fig:gpt_prompt}.

\begin{figure}[t]
\centering
\begin{mdframed}[backgroundcolor=gray!10,roundcorner=6pt]
\ttfamily\footnotesize
\textbf{\underline{System Prompt}}\\
You are an intelligent chatbot designed for evaluating the correctness of 
generative outputs for question--answer pairs. Your task is to compare the 
predicted answer with the correct answer and determine if they match 
meaningfully.  \\

Instructions:
\begin{itemize}
    \item Focus on the meaningful match between the predicted answer and the correct answer.
    \item Consider synonyms or paraphrases as valid matches.
    \item Evaluate the correctness of the prediction compared to the answer.
\end{itemize}

\vspace{4pt}
\textbf{\underline{User Prompt}}\\
Please evaluate the following motion-based question--answer pair:

Question: \{question\}\\
Correct Answer: \{answer\}\\
Predicted Answer: \{pred\}

Provide your evaluation only as a score between 0 and 1, where 1 indicates the 
highest meaningful match. Return the result strictly as a Python dictionary 
string of the form \texttt{\{'score': FLOAT\}}.

Do not add any explanation or additional text.\\

Examples:\\
\texttt{What does the person do?\\
sit vs run --> \{'score': 0.0\}}\\
\texttt{sit vs sit --> \{'score': 1.0\}}\\
\texttt{sit vs sit down --> \{'score': 1.0\}}\\
\texttt{sit vs sit/walk --> \{'score': 0.5\}}\\
\texttt{run vs jog --> \{'score': 1.0\}}\\
\texttt{right hand vs right hand --> \{'score': 1.0\}}\\
\texttt{right hand vs right arm --> \{'score': 0.5\}}\\
\texttt{right hand vs left arm --> \{'score': 0.0\}}
\end{mdframed}
\caption{Prompt template used for GPT-based semantic evaluation on BABEL-QA.}
\label{fig:gpt_prompt}
\end{figure}

\paragraph{ActivityNet-QA Evaluation.}
For ActivityNet-QA, we follow \cite{fang2025humocon, lin2024video} to evaluate generative answers using a GPT-based semantic scoring protocol. Similar to BABEL-QA, we query a GPT model to judge whether a predicted answer is semantically consistent with the ground truth, but we adopt the official ActivityNet-QA scoring scheme in which the model must output both a binary correctness flag and a discrete score in the range $[0,5]$.

All experiments use \texttt{gpt-3.5-turbo-0125} through the batch-completions API. For each sample, we provide two messages: (i) a system prompt specifying the evaluation rules, and (ii) a user prompt containing the question, predicted answer, and ground truth. The GPT model returns a Python dictionary containing \texttt{'pred'} (``yes''/``no'') and \texttt{'score'} (integer in $[0,5]$), which we parse and aggregate following the standard ActivityNet-QA evaluation.
The full prompt used for GPT-based scoring is shown in Fig.~\ref{fig:anet_prompt}.

\begin{figure}[t]
\centering
\begin{mdframed}[backgroundcolor=gray!10,roundcorner=6pt]
\ttfamily\footnotesize

\textbf{\underline{System Prompt}}\\
You are an intelligent chatbot designed to evaluate the correctness of 
generative outputs for question--answer pairs. Compare the predicted answer 
with the correct answer and determine whether they match meaningfully. 

Instructions:
\begin{itemize}
    \item Focus on semantic correctness rather than literal matching.
    \item Accept synonyms and paraphrases when appropriate.
    \item Evaluate the predicted answer and assign a correctness flag and score.
\end{itemize}

\vspace{4pt}
\textbf{\underline{User Prompt}}\\
Please evaluate the following \emph{video-based} question--answer pair:

Question: \{question\}\\
Correct Answer: \{answer\}\\
Predicted Answer: \{pred\}

Return your evaluation strictly as a Python dictionary of the form:\\
\texttt{\{'pred': 'yes'/'no', 'score': INTEGER\}}\\
where \texttt{score} is an integer between 0 and 5.

Do not provide any additional explanation or text.

\end{mdframed}
\caption{Prompt template used for GPT-based semantic evaluation on ActivityNet-QA.}
\label{fig:anet_prompt}
\end{figure}

\subsection{One-Shot Training on NTU-RGB+D 120}
For NTU-RGB+D 120, we follow the one-shot recognition protocol, where only a single labeled example is available for each of the 20 test classes. Because an LLM does not operate with a fixed classification vocabulary, we adapt the zero-shot finetuning procedure following the strategy proposed in prior LLM-based action recognition work~\cite{qu2024llmsgoodactionrecognizers}.

During finetuning, we use the full auxiliary set together with the single example from each test class. For every training sample, we provide the model with a list of 20 randomly selected candidate class names that includes the correct class. The list order is randomly shuffled, and class names are paraphrased to prevent memorization of fixed strings. The model is instructed to predict the index of the correct class in the provided list, along with the corresponding paraphrased class name.

At test time, we supply only the 20 true test classes, again in a randomly shuffled order. The model must identify the correct class solely from this list, without exposure to the unseen class names outside their single example. This setup ensures a controlled one-shot evaluation while enabling the LLM to perform classification despite having no closed output vocabulary.

\subsection{Training Parameters}

This section summarizes the implementation details used across the curriculum training stages. All stages use a cosine learning-rate scheduler with a warm-up ratio of 0.03. Following Video-LLaMA3, we set the maximum token length to 16{,}384.
In Stage 1, the Pose Encoder is initialized with the pretrained robust Pose Encoder from Sec.~3.1, the Pose Translator is randomly initialized, and the LLM is initialized from Video-LLaMA3-7B. Only the Pose Encoder and Pose Translator are optimized in this stage to align them with the LLM feature space; the LLM remains frozen.
For Stages 2--4, we initialize from the previous stage and jointly optimize all components, including the LLM. Once LLM training begins, the learning rates for all modules are reduced. We finetune the LLM using LoRA with rank~64. The global batch size is~128.
Unless otherwise specified, motion data is processed at 30~fps with a maximum
sequence length of 800 frames. CompMo follows its benchmark setting of
20~fps; lower rates are used only for the inference-time subsampling analysis.
Tab.~\ref{tab:train_settings} lists the learning rates for each module across training stages, including the optional finetuning step.

\begin{table}[h]
\caption{Learning rates of the modules in \methodname across the four-stage curriculum and optional finetuning. Stage~1 aligns the Pose Encoder and Translator to the LLM feature space; later stages optimize the full model.}
\label{tab:train_settings}
\centering
\resizebox{\columnwidth}{!}{%
\begin{tabular}{@{}cccc@{}}
\toprule
\multirow{2}{*}{Stage} & \multicolumn{3}{c}{Model} \\ \cmidrule(l){2-4}
 & Pose Encoder & Pose Translator & LLM \\ \midrule
1 & $1.0 \times 10^{-5}$ & $1.0 \times 10^{-3}$ & frozen \\
2 & $2.0 \times 10^{-6}$ & $1.0 \times 10^{-5}$ & $1.0 \times 10^{-5}$ \\
3 & $2.0 \times 10^{-6}$ & $1.0 \times 10^{-5}$ & $1.0 \times 10^{-5}$ \\
4 & $2.0 \times 10^{-6}$ & $1.0 \times 10^{-5}$ & $1.0 \times 10^{-5}$ \\
(finetune) & $2.0 \times 10^{-6}$ & $1.0 \times 10^{-5}$ & $1.0 \times 10^{-5}$ \\ \bottomrule
\end{tabular}%
}
\end{table}

\subsection{Pinhole Projection Details.}
\label{sec:app_pinhole}
For experiments with reprojected 2D input, we convert each 3D motion sequence 
$\mathbf{P}\in\mathbb{R}^{T\times J\times 3}$ into a 2D pose sequence using a fixed pinhole camera. 
For each joint $\mathbf{p}_{t,j}=[X_{t,j},Y_{t,j},Z_{t,j}]^\top$, we first apply a rotation around the $x$-axis with tilt angle $\theta=15^\circ$:
\[
\mathbf{R}_x(\theta)=
\begin{bmatrix}
1 & 0 & 0\\
0 & \cos\theta & -\sin\theta\\
0 & \sin\theta & \cos\theta
\end{bmatrix},
\qquad
\tilde{\mathbf{p}}_{t,j}=\mathbf{R}_x(\theta)\mathbf{p}_{t,j}.
\]
We then translate the rotated pose in depth by a fixed offset $d=4.0$ and clamp the depth for numerical stability:
\[
\tilde{Z}_{t,j}\leftarrow \max(\tilde{Z}_{t,j}+d, 10^{-3}).
\]
Using focal length $f=5.0$ and principal point $(c_x,c_y)=(0,0)$, each joint is projected as
\[
u_{t,j}=f\frac{\tilde{X}_{t,j}}{\tilde{Z}_{t,j}}+c_x,
\qquad
v_{t,j}=f\frac{\tilde{Y}_{t,j}}{\tilde{Z}_{t,j}}+c_y.
\]
Finally, we apply a fixed affine normalization to the projected coordinates:
\[
\begin{bmatrix}
u_{t,j}\\ v_{t,j}
\end{bmatrix}
\leftarrow
0.7\left(
\begin{bmatrix}
u_{t,j}\\ v_{t,j}
\end{bmatrix}
+
\begin{bmatrix}
0.03\\ -0.07
\end{bmatrix}
\right).
\]
The final 2D representation stores the projected coordinates together with a constant confidence channel,
\[
\mathbf{q}_{t,j}=[u_{t,j},v_{t,j},1]^\top,
\]
resulting in $\mathbf{Q}\in\mathbb{R}^{T\times J\times 3}$. 
The third channel is therefore a confidence indicator and does not encode depth. 
All samples use the same synthetic camera parameters rather than sequence-specific camera calibration.

\section{Data Generation}
\label{sec:data_generation}
In this section, we further detail the generation of the data modules we propose in the paper. Tab.~\ref{tab:num_samples} gives an overview of the different data modules, where in our curriculum they are used, and how many samples each module contains. Sec.~\ref{sec:pose_curriculum} further details the pose curriculum, while Sec.~\ref{sec:motion_curriculum} details the motion curriculum.

\begin{table}[h!]
\centering
\caption{Overview of the usage of different data modules over the stages in our curriculum. Additionally, we show the number of samples in each data module. $^*$Indicates we propose the data module ourselves.}
\label{tab:num_samples}
\resizebox{\columnwidth}{!}{%
\begin{tabular}{@{}lccccc@{}}
\toprule
\multirow{2}{*}{Dataset / Module} & \multirow{2}{*}{\makecell{Num.\\Samples}}  & \multicolumn{4}{c}{Stages} \\ \cmidrule(l){3-6} 
 &  & 1 & 2 & 3 & 4 \\ \midrule
PoseScript \cite{delmas2024posescript} & 210k & \checkmark & \checkmark &  &  \\
PS-Short* & 70k & \checkmark & \checkmark &  &  \\
PS-fine* & 280k &  & \checkmark & \checkmark &  \\
PS-QA* & 30k &  & \checkmark & \checkmark &  \\
PoseFix \cite{delmas2024posefixcorrecting3dhuman} & 94k &  &  & \checkmark &  \\
PF-fine* & 181k &  &  & \checkmark &  \\
BABEL-cap* & 27k &  &  & \checkmark & \checkmark \\
HML3D-QA \cite{chen2024motionllm, humanml3d} & 32k &  &  &  & \checkmark \\
FTM-QA* & 64k &  &  &  & \checkmark \\
BABEL-QA \cite{endo2023motion} & 2k &  &  &  & \checkmark \\
VCG+ 112k \cite{maaz2024videogptintegratingimagevideo} & 23k &  &  &  & \checkmark \\ \bottomrule
\end{tabular}%
}
\end{table}

\subsection{Pose Curriculum}
\label{sec:pose_curriculum}
\subsubsection{LLM Re-Annotation \& Judging}
\paragraph{PS-Short.} To construct PS-Short, we convert the long PoseScript \cite{delmas2024posescript} descriptions into concise one-sentence summaries. For each pose, we use Qwen2.5-VL-7B-Instruct, conditioning the model on both the rendered SMPL image and two of the original PoseScript descriptions. The model produces a short and accurate summary without referencing the image or hallucinating additional details.

To ensure quality, every generated summary is then judged using a separate prompt. The same model receives the ground-truth description together with multiple candidate summaries (one correct, several distractors) and selects which one best matches the pose semantics. Only summaries judged correct are retained.

Fig.~\ref{fig:ps_short_prompts} shows the exact prompt templates used for both generation and judging.

\begin{figure}[h!]
\centering
\begin{mdframed}[backgroundcolor=gray!10,roundcorner=6pt]
\ttfamily\footnotesize

\textbf{\underline{Generation Prompt}}\\[2pt]
Describe the human pose in this image accurately and concisely in one sentence. Use the following two extensive descriptions of the pose to help you understand it better, but do not include any extra information that is not in the descriptions. Do not mention anything about SMPL or image in the summary.\\[4pt]

\texttt{Image: <SMPL .png render>\\
Descriptions: \\
1. <desc\_1>\\
2. <desc\_2>\\
}  
\rule{\linewidth}{0.3pt}

Expected output:

\texttt{<A concise one-sentence summary of the pose.>}

\vspace{10pt}

\textbf{\underline{Judging Prompt}}\\[2pt]
You receive a description of a human pose or activity.\\
Below are five summaries of the description, one of which is the correct summary. Only respond with the corresponding multiple-choice letter.\\[4pt]

\texttt{
Description: <desc\_3>\\[2pt]
A) \{summary\_1\}\\
B) \{summary\_2\}\\
C) \{summary\_3\}\\
D) \{summary\_4\}\\
E) \{summary\_5\}\\
} 
\rule{\linewidth}{0.3pt}
Expected output:  
\texttt{C}

\end{mdframed}

\caption{Prompt templates used to construct PS-Short. 
Top: generation prompt used by Qwen2.5-VL to produce concise one-sentence summaries from SMPL renders and PoseScript descriptions.
Bottom: judging prompt used to automatically filter incorrect summaries.}
\label{fig:ps_short_prompts}
\end{figure}

\paragraph{PS-fine.}
To construct PS-fine, we target fine-grained pose understanding by extracting body-part–level motion descriptions from the original PoseScript \cite{delmas2024posescript} data. Unlike PS-short, which summarizes an entire pose, PS-fine captures \emph{local} motion semantics: for each pose transition description, we ask LLaMA-3.3-70B-Instruct to (i) select two body parts mentioned in the description, (ii) generate a short and precise sentence describing the movement of each selected body part, and (iii) produce an `opposite' version of the pose by minimally altering attributes such as location or configuration. These structured JSON outputs yield compact, fine-grained pose primitives suitable for training the LLM on localized pose reasoning.

As in PS-Short, all generated descriptions undergo automatic quality control. For each pose sample, we construct a multiple-choice prompt with four options: constructed out of the correct and 'opposite' descriptions of the two body parts. The LLM receives another full PoseScript description together with the four candidate options and selects the correct one. Only body-part descriptions passing this verification step are retained. Fig.~\ref{fig:ps_fine_prompts} presents the prompt templates used for PS-Fine generation and automatic judging.

\begin{figure}[h!]
\centering
\begin{mdframed}[backgroundcolor=gray!10,roundcorner=6pt]
\ttfamily\footnotesize

\textbf{\underline{Generation Prompt}}\\[2pt]
You are given a description of a pose:\\
\texttt{<desc\_1>}\\[4pt]

Task:
\begin{itemize}\itemsep2pt
    \item Pick two body parts that are described in the description.
    \item Describe the pose for the chosen body part, always include the body part in the sentence but don't describe other body parts.
    \item Describe the opposite of the pose, only alter the specific difference (i.e. higher instead of lower, straight instead of bent).\\[2pt]
\end{itemize}

Respond only with two JSONs. Do not include explanations, reasoning, or repeat the prompt.

Output format: \\
\{"body\_part": "<chosen body part>", \\
"description": "<one short sentence>", \\ 
"opposite": "<one short sentence>"\} \\
\{"body\_part": "<chosen body part>", \\
"description": "<one short sentence>", \\ 
"opposite": "<one short sentence>"\} \\
\rule{\linewidth}{0.4pt}

\textbf{\underline{Judging Prompt}}\\[2pt]
You are given a description of a pose:\\
\texttt{<desc\_2>}\\[2pt]

Below are four pose descriptions of \texttt{<body\_part1>} and \texttt{<body\_part2>}.
Pick the correct option.  
Only return a JSON dictionary of the form:\\[2pt]
\texttt{\{"choice": "a"\}}\\[6pt]

\texttt{A)<body\_part1 correct> <body\_part2 correct>\\
B)<body\_part1 correct> <body\_part2 opposite>\\
C)<body\_part1 opposite> <body\_part2 correct>\\
D)<body\_part1 opposite> <body\_part2 opposite>\\
}

\end{mdframed}

\caption{Prompt templates used for creating PS-Fine. Top: fine-grained generation prompt used to extract body-part–level motion descriptions and their opposites from PoseScript descriptions using LLaMA-3.3-70B. Bottom: judging prompt used to automatically filter incorrect or inconsistent descriptions through multiple-choice verification.}
\label{fig:ps_fine_prompts}
\end{figure}

\paragraph{PF-fine.}
To construct PF-fine, we follow a procedure analogous to PS-fine but applied to the PoseFix \cite{delmas2024posefixcorrecting3dhuman} dataset, which provides motion descriptions that express how to transition from pose A to pose B. Our goal is to convert these global transition descriptions into fine-grained body-part–level motion primitives that teach the LLM localized motion understanding.

For each transition description, we prompt LLaMA-3.3-70B-Instruct to:  
(i) select two body parts explicitly mentioned or implied in the sentence,  
(ii) generate a one-sentence description of how each selected body part moves from pose A to pose B, and  
(iii) generate an 'opposite' motion description by minimally flipping the movement direction (e.g., forward→backward, lifted→lowered).  
The model returns two JSON objects, each containing a body part, its fine-grained motion description, and its opposite. These pairs form the PF-fine motion primitives.

As with PS-fine, we apply automatic verification by constructing a multiple-choice judging prompt. Each question contains four options formed by systematically combining correct and 'opposite' motion descriptions across the two body parts. Given the original PoseFix transition description, the LLM must select the correct combined option. Only motion primitives that pass this consistency check are retained. 
Fig.~\ref{fig:pf_fine_prompts} shows the generation and judging prompts used for constructing PF-fine.

\begin{figure}[h!]
\centering
\begin{mdframed}[backgroundcolor=gray!10,roundcorner=6pt]
\ttfamily\footnotesize

\textbf{\underline{Generation Prompt}}\\[2pt]
You are given a description of how to move from pose A to pose B:\\
\texttt{<desc\_1>}\\[4pt]

Task:
\begin{itemize}\itemsep2pt
    \item Pick two body parts that are described in the transition.
    \item Describe the movement for each chosen body part; always include the body part in the sentence.
    \item Describe the opposite of the movement by altering only the specific difference (e.g., backward instead of forward).\\[2pt]
\end{itemize}

Respond only with two JSONs. Do not include explanations, reasoning, or repeat the prompt.

Output format:\\
\{"body\_part": "<chosen body part>",\\
 "description": "<one short sentence>",\\
 "opposite": "<one short sentence>"\}\\
\{"body\_part": "<chosen body part>",\\
 "description": "<one short sentence>",\\
 "opposite": "<one short sentence>"\}\\

\rule{\linewidth}{0.4pt}

\textbf{\underline{Judging Prompt}}\\[2pt]
You are given a description of moving from pose A to pose B:\\
\texttt{<desc\_2>}\\[4pt]

Below are four options describing the movements of \texttt{<body\_part1>} and \texttt{<body\_part2>}.\\
Pick the correct option.  
Only return a JSON dictionary of the form:\\[2pt]
\texttt{\{"choice": "a"\}}\\[6pt]

\texttt{A)<body\_part1 correct> <body\_part2 correct>\\
B)<body\_part1 correct> <body\_part2 opposite>\\
C)<body\_part1 opposite> <body\_part2 correct>\\
D)<body\_part1 opposite> <body\_part2 opposite>\\
}

\end{mdframed}

\caption{Prompt templates used for creating PF-Fine. 
Top: generation prompt used to extract body-part–level motion descriptions and their opposites from PoseFix transition descriptions using LLaMA-3.3-70B. 
Bottom: judging prompt used to verify motion correctness through multiple-choice consistency checking.}
\label{fig:pf_fine_prompts}
\end{figure}

\subsubsection{PoseScript Question \& Answer}
\label{sec:app_ps-qa}
We further introduce PS-QA, a machine-generated dataset that produces structured question--answer pairs directly from 3D pose geometry.  
Unlike PS-Short and PS-Fine, PS-QA does not rely on LLM re-annotation: every answer is computed analytically from joint coordinates.  
The goal of PS-QA is to introduce the ability to reason about spatial relations, and identify geometric patterns such as symmetry or joint configuration.
PS-QA is built using six complementary question families, each targeting a different aspect of spatial reasoning:

\begin{itemize}\itemsep2pt
    \item \textbf{Closest Joint.}  
    Given a starting joint (e.g., wrist or ankle), the model must identify which other joint is spatially closest while excluding the joint’s direct parent in the kinematic chain.  
    We compute Euclidean distances in 3D and ensure that closest candidates are well-separated to avoid ambiguity.

    \item \textbf{Furthest Joint.}  
    Similar to the above but requiring the identification of the most distant joint relative to a starting joint.  
    Joint distributions are balanced using frequency constraints to avoid bias toward particular anatomical regions.

    \item \textbf{Feet Relation.}  
    This category determines whether the left or right foot is in front or behind.  
    We estimate body orientation using the hip–spine plane and compare the projected foot positions along the forward axis.  
    Only poses with a sufficiently clear front/back difference are included.

    \item \textbf{Symmetry Relation.}  
    We evaluate symmetry by mirroring the left-side joints across the inferred sagittal plane and computing the left--right joint deviations.  
    Depending on the magnitude of these deviations, we label the pose as having \emph{full}, \emph{upper-body}, \emph{lower-body}, or \emph{no} symmetry.  
    This produces categorical answers that test higher-level structural understanding.

    \item \textbf{Joint Bending.}  
    For joints such as the knees, elbows, hips, and spine, we compute bending angles using the standard angle between limb vectors.  
    Angles are discretized into interpretable linguistic categories following~\cite{delmas2024posescript} (\emph{straight}, \emph{slightly bent}, \emph{partially bent}, \emph{right angle}, \emph{almost completely bent}, \emph{completely bent}).  
    Each category is evenly represented.

    \item \textbf{Comparative Limb Bending.}  
    The model compares the bending of two joints (e.g., left vs.\ right knee) and decides which is more (or less) bent.  
    Only pairs with a sufficiently large angle difference are retained to avoid uncertain cases.
\end{itemize}

All questions are phrased using multiple natural-language templates, and the ground-truth answer strings are diversified using paraphrases, while remaining strictly determined by geometry.  
Each instance additionally includes a small set of multiple-choice alternatives, enabling training and evaluation of both free-form and forced-choice reasoning.

Table~\ref{tab:psqa_categories} summarizes the PS-QA question types and their geometric computations.

\begin{table}[h!]
\centering
\footnotesize
\begin{tabular}{p{2.7cm} p{4.8cm}}
\toprule
\textbf{Category} & \textbf{Underlying Computation} \\
\midrule
Closest Joint & 3D Euclidean distance to a starting joint, parent excluded \\
Furthest Joint & Maximum 3D distance from starting joint \\
Feet Relation & Front/back ordering via hip--spine plane projection \\
Symmetry Relation & Left--right mirrored joint MPJPE thresholds \\
Joint Bending & Limb angle calculation and discretization \\
Compare Limb Bending & Relative angle difference between two joints \\
\bottomrule
\end{tabular}
\caption{Overview of PS-QA question categories and their geometric computation rules.}
\label{tab:psqa_categories}
\end{table}

\subsection{Motion Curriculum}
We further detail BABEL-cap, FTM-QA, and HML3D-QA. Across all constructed pose- and motion-level supervision modules, we exclude source motion sequences belonging to the validation or test splits of the evaluated benchmarks, including BABEL-QA and CompMo. This exclusion also covers individual poses and clips derived from those sequences. In particular, because BABEL-QA uses a train/val/test split that does not align with the original BABEL split, we remove from the BABEL-cap and FTM-QA train/val sets all motion IDs appearing in the BABEL-QA validation or test splits.
\label{sec:motion_curriculum}
\subsubsection{BABEL-cap}
To obtain simple motion captions for the early stage of our curriculum, we construct BABEL-cap, a set of short AMASS~\cite{mahmood2019amass} clips paired with concise BABEL~\cite{punnakkal2021babel} action labels. Each clip is linked to a brief question–answer pair (e.g., \emph{“What action is being performed?”}).

We extract raw action labels from BABEL’s \texttt{frame\_ann} entries when available, otherwise falling back to \texttt{seq\_ann}. We discard labels that do not describe a meaningful atomic action for short clips, including:
\begin{itemize}
    \item non-informative labels (\texttt{transition}, \texttt{unknown});
    \item context-dependent labels (e.g., ``walk back to'', ``back to original position'');
    \item segments shorter than two frames.
\end{itemize}

For each valid action segment, we form a short prompt requesting a minimal action description and pair it with the corresponding BABEL label. We use several interchangeable prompt phrasings (e.g., ``Describe the action in a few words.''), but do not rely on any complex template logic. Each sample specifies the referenced AMASS motion file, the prompt, the answer, and the annotated time span.

This results in a compact set of atomic action descriptions that serve as the first stage of our motion curriculum.

\subsubsection{FTM-QA}
\label{app:ftmqa}

FTM-QA extends the temporal supervision available in BABEL by transforming its sequence- and frame-level annotations into structured question--answer pairs. For each motion, we extract the global sequence label together with all frame annotations that include explicit start and end times, ordering them chronologically and discarding non-semantic labels such as \texttt{transition} and \texttt{unknown}. The remaining labels are formatted into a single annotation block that lists the overall activity followed by each temporally grounded segment. This block forms the sole evidence the model may rely on.

To construct QA pairs, we follow the re-annotation strategy of \cite{chen2024motionllm} while adapting it to the richer temporal structure of BABEL. The LLM, Qwen2.5-72B-Instruct, is instructed to generate only questions whose answers can be unambiguously inferred from the motion sequence, without referencing unseen descriptions, relying on the numbering of annotations, or assuming information not encoded in the temporal labels. The prompt highlights typical forms of temporal reasoning—identifying start or end times, determining what action follows or precedes another, relating short events to the global activity, and describing changes over time—while avoiding ambiguity or hallucination. The model returns a JSON list containing the generated question--answer pairs, which we then directly associate with the corresponding AMASS \cite{mahmood2019amass} motion clip. This produces a dataset explicitly aligned with temporal ordering, duration reasoning, and segment-level structure, complementing the spatial supervision provided by PS-QA.

Fig.~\ref{fig:ftm_qa_prompt} shows the prompt template used for re-annotation.

\begin{figure*}[t]
\centering
\begin{mdframed}[backgroundcolor=gray!10,roundcorner=6pt]
\ttfamily\scriptsize

This is the complete annotation of one motion sequence. Sequence label contains information about the whole sequence, while frame labels contain information about specific segments indicated by corresponding start-time and end-time. Sequence labels can give context to frame labels. Please construct several QA pairs based on this information.

! Note that the sequence label does not have a start or end time.\\
! Note that each frame label has its own start and end time, one label can have to do with a previous one (i.e. 'move back to the original position').\\
! Note that don't ask questions about the frame annotations: 'transition' and 'unknown'.\\
! Note that you can only see the movement, not the descriptions in advance. Therefore, you can NOT ask or answer something like 'the first/third description'.\\
! Note that you should only propose questions for which you are sure they are possible to answer using the motion sequence. Avoid ambiguity.\\
! Note that Don't say anything like 'sure' or 'here is xxx', just return the QA's directly in the form of a JSON.\\
Don't rigidly imitate the template either.\\[8pt]

\text{HERE IS AN EXAMPLE:}\\[2pt]
\text{[GIVEN DESCRIPTION]:}\\
Sequence label:\\
throwing a baseball\\
Frame labels:\\
Stand \#0.0--0.4\\
Transition \#0.4--0.82\\
Throw ball with left hand \#0.8--2.1\\
Transition \#2.1--2.8\\
Retreat right foot \#2.8--3.7\\
Stand \#3.7--5.0\\
Walk to left \#5.0--7.0\\[6pt]

\text{[System output]:}\\
\text{[}
  \{"q": "What is the main action performed in this sequence?",
   "a": "The person throws a baseball."\},\\
  \{"q": "What does the person do after throwing the ball?",
   "a": "They retreat their right foot and return to a standing position."\},\\
  \{"q": "At what point does the throwing action begin?",
   "a": "Around 0.8 seconds."\},\\
  \{"q": "What happens before walking away?",
   "a": "The person stands for a moment after retreating their right foot."\},\\
  \{"q": "Which hand is used for the throw?",
   "a": "The left hand."\},\\
  \{"q": "Does the person walk away immediately after throwing?",
   "a": "No, the person stands for a good second before walking away."\},\\
  \{"q": "Does the person walk away, if so in what direction?",
   "a": "Yes, the person walks away to the left."\}\\
\text{]} \\[10pt]

\text{ACTUAL SAMPLE:}\\
\texttt{<sample\_annotations>}

\end{mdframed}

\caption{Prompt template used to generate FTM-QA. The LLM receives sequence-level and frame-level BABEL annotations and produces temporally grounded question–answer pairs in JSON format.}
\label{fig:ftm_qa_prompt}
\end{figure*}

\subsubsection{HML3D-QA}
Following prior work \cite{chen2024motionllm, fang2025humocon, li2025human}, we additionally construct question--answer pairs for HumanML3D using an LLM-based procedure. Each HumanML3D sample provides several textual descriptions of the same motion.
We follow the established pipeline by supplying these multi-description annotations to an LLM, we use Qwen2.5--72B-Instruct, and prompting it to generate diverse motion QA pairs grounded in the motion itself. The LLM is instructed to: (i) avoid referencing the description text explicitly, (ii) ensure all questions are answerable solely by observing the motion sequence, and (iii) avoid ambiguous or overly speculative questions. The LLM returns a JSON list of QA pairs, which we store directly as the HumanML3D-QA dataset.

Fig.~\ref{fig:hml3d_prompt} shows the prompt template used for re-annotation.

\begin{figure*}[t]
\centering
\begin{mdframed}[backgroundcolor=gray!10,roundcorner=6pt]
\ttfamily\scriptsize

These are multiple descriptions of the same motion sequence, each line contains a different description of the motion. The last two numbers in each line correspond to the start-time and end-time of the description. If the numbers are all 0.0, it represents the entire sequence. Please construct several motion QA pairs based on this information, the question should be answerable by only viewing the motion.

! Note that the start and end time of each line only correspond to the description of that line, and has nothing to do with the other lines. Avoid ambiguity.\\
! Note that only the movement is provided with the question, NOT the descriptions. Therefore, you can NOT ask or answer something like 'the first/third description' or 'in the description'.\\
! Note that you should not mention anything about the descriptions in the QA's. Instead of saying: 'it is described as big' say 'it is big'.\\
! Note that you should only propose questions for which you are sure they are possible to answer using the motion sequence. Avoid ambiguity.\\
! Note that Don't say anything like 'sure' or 'here is xxx', just return the QA's directly in the form of a JSON.\\
Don't rigidly imitate the template either.\\[6pt]

\text{HERE IS AN EXAMPLE:}\\
\text{[GIVEN DESCRIPTION]:}\\
a person walks toward the front, turns to the right, bounces into a squat, and places both arms in front of chest before placing them on the knees.\#0.0\#5.0\\
person walks up and squats slightly to pose a position\#0.0\#0.0\\
he moved forward then stretched his body, moving his hand and touching his knees with hands and now he turned to the left side.\#0.0\#0.0\\
a person walks, turns slightly to the right, squats, puts hand on both knees while squatting, and then squats again.\#0.0\#8.0\\[4pt]

\text{[System output]:}\\
\text{[}
\{"q": "What is the motivation behind the person performing these motions?",
 "a": "Without additional context, it is not possible to determine the exact motivation."\},\\
\{"q": "True or False: The person touches their knees with their hands immediately after walking.",
 "a": "False. He stretches his body before touching his knees."\},\\
\{"q": "What does the man do after walking forward?",
 "a": "The guy turns to the right."\},\\
\{"q": "What does the guy do after walking forward and turning right?",
 "a": "He bounces into a squat."\},\\
\{"q": "How many times does the man squat?",
 "a": "twice."\},\\
\{"q": "Can you describe the initial movement of the person in this sequence?",
 "a": "Initially, the person begins by walking forward."\},\\
\{"q": "Could you elaborate on the arm movements of the person during the squat?",
 "a": "They place both arms in front of the chest and then on the knees."\}
\text{]}\\[8pt]

\text{ACTUAL SAMPLE:}\\
\texttt{<sample\_annotations>}

\end{mdframed}
\caption{Prompt template used to generate HML3D-QA. The LLM is given multiple natural-language descriptions of the same motion and asked to produce a JSON list of motion-grounded QA pairs.}
\label{fig:hml3d_prompt}
\end{figure*}

\section{Limitations and Broader Impact}
\label{sec:limitations_impact}

\paragraph{Limitations.}
Although \methodname supports both 2D and 3D skeletal input, its performance depends on the quality of the input poses. Noisy detections, missing joints, severe occlusion, or unusual camera viewpoints may reduce performance. Long sequences are handled through uniform temporal subsampling, which may miss short actions; adaptive temporal sampling is a promising direction for future work. Finally, while we evaluate temporal localization on CompMo and motion QA on multiple benchmarks, broader real-world deployment would require testing across more diverse environments, activities, camera settings, and populations.

\paragraph{Broader impact.}
Skeletal motion representations can support privacy-preserving applications such as rehabilitation monitoring, sports coaching, and human-robot interaction by reducing reliance on raw video. However, motion-understanding systems could also be misused for surveillance or behavioral monitoring if deployed without consent or appropriate safeguards. Since pose estimators and motion datasets may have demographic or viewpoint biases, downstream systems should be evaluated carefully before deployment in sensitive settings.

\section{Compute Resources}
\label{sec:compute}

All experiments were run on 4 NVIDIA A100 GPUs with 80 GB memory per GPU. The main \methodname training required approximately 150 GPU-hours, and each downstream fine-tuning run required approximately 4 GPU-hours. The total compute used for the reported experiments was approximately 478 GPU-hours. Additional preliminary experiments were conducted during development but are not included in this estimate.

\section{Assets, Licenses, and Release}
\label{sec:assets}

We use existing datasets and models including AMASS, Human3.6M, PoseTrack, InstaVariety, PoseScript, PoseFix, BABEL, HumanML3D, HuMMan-QA, CompMo, NTU-RGB+D 120, QEVD-Coach, VideoLLaMA3, Qwen2.5, and MMPose. We cite the original sources throughout the paper and use them according to their respective licenses and terms of use. Our derived supervision modules are constructed from these existing resources and will be released together with code and documentation, subject to the licenses and redistribution terms of the underlying datasets. The released documentation will include data construction procedures, prompt templates, filtering details, and training/evaluation instructions.

\end{document}